%% file: arxiv.tex
\documentclass[10pt,twocolumn,letterpaper]{article}

\usepackage{cvpr}              
\usepackage{multirow}
\usepackage{amssymb}
\usepackage{bbding}
\usepackage{overpic}
\usepackage[accsupp]{axessibility}
\usepackage{float}

\usepackage{graphicx}

\usepackage[normalem]{ulem}
\usepackage{tabularray}
\newcommand{\minisection}[1]{\vspace{0.04in} \noindent {\bf #1}\ \ }

\input{preamble}

\definecolor{cvprblue}{rgb}{0.21,0.49,0.74}
\usepackage[pagebackref,breaklinks,colorlinks,citecolor=cvprblue]{hyperref}

\def\paperID{4647} 
\def\confName{CVPR}
\def\confYear{2024}

\title{Generative Multi-modal Models are Good Class-Incremental Learners}

\author{Xusheng Cao$^1$, Haori Lu$^1$, Linlan Huang$^1$, Xialei Liu$^{2,1}$\thanks{Corresposing author.}, Ming-Ming Cheng$^{2,1}$\\
$^1$VCIP, CS, Nankai University \qquad $^2$NKIARI, Shenzhen Futian\\
{\tt\small \{caoxusheng, luhaori, huanglinlan\}@mail.nankai.edu.cn, \{xialei, cmm\}@nankai.edu.cn}
}

\begin{document}
\maketitle
\input{sec/0_abstract}    

\input{sec/1_intro}
\input{sec/2_relatedwork}

\input{sec/3_method}

\input{sec/4_experiments}

\section{Conclusion}
In this paper, we propose GMM to use generative models for class-incremental learning. By fine-tuning the Generative Multi-modal Model (GMM), we directly generate the label text of the images to be classified. Then we select the label most similar to the generated text by its features. Our experiments demonstrate that this method, which does not require a classification head, is highly effective in addressing classification biases in continual learning.

\minisection{Limitations. } 
Since we are the first to introduce generative models to class-incremental learning, the overall design of our method is embarrassingly simple. We believe that with more focused efforts in this direction, there will be significant advancements in the field of continual learning.

\minisection{Broader impact. } 
We believe that introducing GMM into continual learning (CL) is both necessary and urgent. With the rapid development of GMM, we can leverage their capabilities to improve the performance of continual learning. Besides, integrating CL methods into the training process of GMM could significantly reduce training costs.

\minisection{Acknowledgments.}
This work is funded by  
NSFC (NO. 62206135, 62225604), Young Elite Scientists Sponsorship Program by CAST (2023QNRC001), and the Fundamental Research Funds for the Central Universities 
(Nankai Universitiy, 070-63233085). 
Computation is supported by the Supercomputing Center of Nankai University.

{
    \small
    \bibliographystyle{ieeenat_fullname}
    \bibliography{main}
}

\end{document}

%% file: preamble.tex
\usepackage[dvipsnames]{xcolor}


%% file: sec/0_abstract.tex
\begin{abstract}
In class-incremental learning (CIL) scenarios, the phenomenon of catastrophic forgetting caused by the classifier's bias towards the current task has long posed a significant challenge. It is mainly caused by the characteristic of discriminative models. With the growing popularity of the generative multi-modal models, we would explore replacing discriminative models with generative ones for CIL. However, transitioning from discriminative to generative models requires addressing two key challenges. The primary challenge lies in transferring the generated textual information into the classification of distinct categories. Additionally, it requires formulating the task of CIL within a generative framework. To this end, we propose a novel generative multi-modal model (GMM) framework for class-incremental learning. Our approach directly generates labels for images using an adapted generative model. After obtaining the detailed text, we use a text encoder to extract text features and employ feature matching to determine the most similar label as the classification prediction. In the conventional CIL settings, we achieve significantly better results in long-sequence task scenarios. Under the Few-shot CIL setting, we have improved by at least 14\% accuracy over all the current state-of-the-art methods with significantly less forgetting. Our code is available at \url{https://github.com/DoubleClass/GMM}.
\end{abstract}

%% file: sec/1_intro.tex
\section{Introduction}
\label{sec:intro}

Deep neural networks~\cite{liu2023survey,guo2023visual, sun2023corrmatch} have made remarkable strides in numerous applications, primarily owing to the vast amounts of data and computational resources at their disposal. Nonetheless, these accomplishments are predominantly contingent on having access to all the required data simultaneously for training on various tasks. In cases where data is acquired incrementally, these networks often encounter the challenge of catastrophic forgetting~\cite{mccloskey1989catastrophic}. Hence, the capacity to seamlessly incorporate new knowledge while retaining previously acquired knowledge is a highly desirable attribute for future artificial intelligence systems. Continual learning~\cite{qu2021recent,wang2023comprehensive, yang2023continual,zhou2023deep} is a subject of study aimed at advancing the evolution of neural networks toward this goal.

Numerous studies have delved into continual learning, categorizing their approaches into three main groups~\cite{delange2021continual}: rehearsal-based, architecture-based, and regularization-based methods. Additionally, hybrid methods are gaining popularity as they combine insights from different perspectives. Within this research landscape, three primary scenarios~\cite{van2019three} have received extensive attention, with class-incremental learning (CIL)~\cite{masana2022class} being one of the most demanding settings. In our work, we concentrate on CIL, wherein each task comprises a distinct set of classes, and the primary challenge is to enable the network to recognize new classes without forgetting knowledge of previously encountered ones.

\begin{figure}
\centerline{\includegraphics[width=0.47\textwidth]{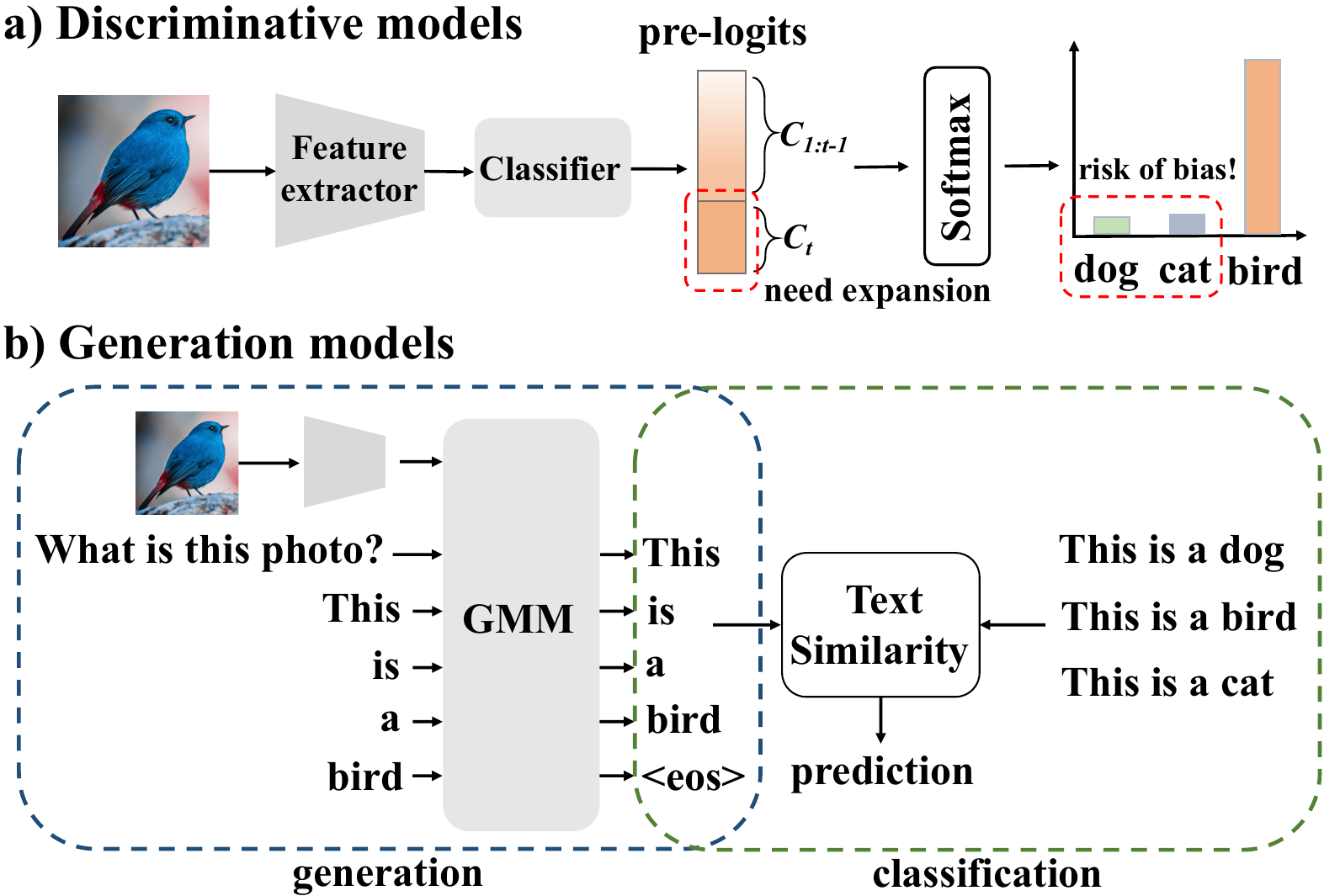}}
\caption{Illustration of conventional discriminative models for class-incremental learning (CIL) and our generative multi-modal models (GMM) for CIL. Discriminative models pose potential risk of classifier bias toward current task with network expasion. Our GMM framework consists generation and classification phases. And It is adapted to CIL based on the similarity of generated text and the true category names. }

\label{fig:intro}
\end{figure}

The majority of existing research on class-incremental learning (CIL) focuses on training models from scratch, relying solely on data from the current tasks~\cite{ BiC, castro2018end, douillard2020podnet, douillard2022dytox, kirkpatrick2017overcoming, li2017learning,rebuffi2017icarl,UCIR_2019_CVPR,yan2021dynamically,   zhu2021prototype}. In contrast, humans accumulate knowledge over extended periods, drawing upon a wealth of prior world knowledge. Consequently, there has been a growing interest in pre-trained models for CIL~\cite{l2p, smith2023coda, wang2022dualprompt,zhang2023slca,zhou2023learning}, harnessing the knowledge acquired from extensive, pre-existing datasets to address the immediate tasks at hand. For instance, prompt-based methods~\cite{l2p, smith2023coda, wang2022dualprompt} utilize prompt-tuning to summrize task-specific knowledge from prior pre-trained knowledge. SLCA~\cite{zhang2023slca} and ADAM~\cite{zhou2023revisiting}, on the other hand, solely fine-tuning the pre-trained model to adapt the pre-existing knowledge into the objectives of immediate tasks.

To tackle image classification downstream tasks, pre-trained models are traditionally derived from discriminative tasks, like supervised learning on datasets such as ImageNet-21K~\cite{russakovsky2015imagenet}, or they may stem from self-supervised learning efforts~\cite{caron2021emerging, chen2020improved, chen2021mocov3, he2020momentum}. However, in our research, we venture into the paradigm of generative multi-modal models to address image classification tasks. Generative models like GPT4~\cite{openai2023gpt4} and LLaVa~\cite{liu2023visual} have garnered significant attention in recent years due to their capacity to produce highly informative descriptions of input images. On the one hand, it can harness the wealth of semantic correspondences between texts and images, while on the other hand, there's no requirement to expand the classifier with each new task, unlike in the case of discriminative models for CIL. 

Nonetheless, harnessing the knowledge from pre-trained generative models for downstream class-incremental learning (CIL) tasks presents a nontrivial endeavor. The primary challenge lies in transferring the generated textual information into the classification of distinct categories. Additionally, there's the task of formulating CIL within a generative framework, which poses a second significant challenge. Shao et al.~\cite{shao2023class} presents the VAG system, which formulates CIL as a continual label generation problem, preserving the language model's ability to learn new classes.
However, it only works in the field of Natural Language Processing (NLP), which is inherently suited to Large Language Models (LLM). To the best of our knowledge, we are the first to apply this generative approach to incremental learning in the field of image classification.

In this work, we propose Generative Multi-modal Models (GMM) for class-incremental learning. As illustrated in Fig.~\ref{fig:intro} (a), conventional discriminative methods extract image features with a network backbone, then forward them to a classifier to obtain the probability of the image belonging to each label, with the label having the highest probability being the output of the discriminative models.  While in Fig.~\ref{fig:intro} (b), we adopt a generative approach, which directly produces a descriptive sentence for the given image, which is then compared with the actual label texts with a text encoder. The most similar label becomes the predicted result of our generative model.    
This approach allows us to leverage the rich pre-training knowledge in generative multi-modal models while avoiding the use of the expanded classification head, which mitigates the risk of the model bias towards the current task and reduces catastrophic forgetting. 

The main contributions of this paper are:

\begin{itemize}
    \item  We propose a novel generative approach (GMM) to address class-incremental learning by leveraging multi-modal models.
    \item We reformulate GMM for image classification and adapt it for the downstream benchmarks.  Without an expanded classification head like in discriminative models, our model significantly mitigates the issue of bias towards current tasks, resulting in significantly reduced forgetting in CIL.
    \item Our model achieves state-of-the-art performance across multiple datasets in both conventional and few-shot CIL settings.
\end{itemize}

%% file: sec/2_relatedwork.tex
\section{Related Work}
\label{sec:work}

\subsection{Class-Incremental Learning}
In Class-Incremental Learning, tasks arrive sequentially, and each class is exclusive to a specific task without any overlap. The goal is to acquire knowledge from new classes while preserving information from previously encountered classes.
There are three primary branches in CIL~\cite{delange2021continual}, including rehearsal-based, architecture-based, and regularization-based methods.
Rehearsal-based methods~\cite{ahn2021ss,castro2018end,rebuffi2017icarl,wu2019large} store a small set of data derived from old classes to represent knowledge from previous tasks. 
These exemplar data can be either original data~\cite{rebuffi2017icarl}, generative data~\cite{gao2023ddgr, shin2017continual} or hidden features~\cite{hayes2020remind}. 
Architecture-based methods focus on modifying network architecture to alleviate forgetting.
Approaches include learning redundant network architecture~\cite{fernando2017pathnet, rajasegaran2019random}, learning different expert networks~\cite{aljundi2017expert,schwarz2018progress} or parameters~\cite{liu2021adaptive, mallya2018piggyback,  serra2018overcoming} for each task, dynamically expanding network parameters to accumulate incremental knowledge~\cite{yan2021dynamically}. 
Regularization-based methods introduce an additional regularization term to restrict network updates when adapting to new tasks. In such cases, EWC~\cite{kirkpatrick2017overcoming}, SDC~\cite{yu2020semantic} and Rotated-EWC~\cite{liu2018rotate} expect that parameters essential for the old tasks should not be updated excessively. 
Moreover, from the perspective of network output consistency, numerous studies~\cite{hu2021distilling,li2018learning,liu2022long,tao2020topology,zhang2022representation} incorporate distillation to prevent forgetting.

\minisection{Few-shot CIL}
Few-shot Class-Incremental Learning (FSCIL)~\cite{mazumder2021few, tao2020few}  explores few-shot learning in an incremental context, with all data samples available for base session and very limited data in each incremental session. 
Some FSCIL methods~\cite{cheraghian2021semantic,tao2020few,zhao2021mgsvf} train the model in both base and incremental sessions, aiming to mitigate overfitting challenges caused by the limited data in incremental learning. Other strategies~\cite{shi2021overcoming,zhang2021few,zhu2021self} primarily train the model in the base session and make minimal adjustments in the incremental sessions, thereby reducing forgetting but may come at the cost of decreased precision in the incremental sessions.

\subsection{Pre-trained models for CIL}
There are many methods~\cite{continual-clip,l2p,wu2022class} having shown that pre-trained models are effective for continual learning. 
One main branch trains a set of prompts to retain previous knowledge~\cite{l2p, smith2023coda,wang2022dualprompt,wang2024hierarchical}. A selected subset of prompts are fed into the model during forward to prompt model the past knowledge. 
Additionally, methods like SLCA~\cite{zhang2023slca} and ADAM~\cite{zhou2023revisiting} fine-tune pre-trained models, achieving impressive results with less forgetting. 
Continual-CLIP~\cite{continual-clip} demonstrates that the CLIP~\cite{clip} model is capable of performing continual learning without any extra training. This highlights the significant potential of multi-modal pre-trained models in the realm of continual learning.
Inspired by this, many methods~\cite{zhou2023learning, liu2023class} employ CLIP as the backbone to utilize the multi-modal information. However, if not using a classifier directly, these approaches need to utilize expanded text features to calculate distances with image features for classification. This will exacerbate the model's bias towards current data, consequently leading to the forgetting of previously acquired knowledge. To avoid this bias, we use a generative model to directly generate prediction text. The fixed text decoder will function as the classifier, significantly alleviating the bias.

\subsection{Vision Language Models}
In recent years, vision-language multi-modal models have made significant progress and achieved impressive results in various downstream tasks~\cite{ben2017mutan,li2019object,liu2021cptr,wang2023large}. 
Traditional vision-language models employ different types of encoders to extract information from vision and language models, including single-stream~\cite{sun2019videobert}, dual-stream~\cite{lu2019vilbert} and fusion~\cite{tan2019lxmert} encoders.
A key aspect of vision-language models is the alignment of multi-modal features. 
CLIP~\cite{clip}, for example, extracts image and text features separately using respective encoders and enforces alignment through a contrastive loss, ensuring alignment between positive image-text pairs in the feature space.
VisualGPT~\cite{chen2022visualgpt} and Frozen~\cite{tsimpoukelli2021multimodal} leverage pre-trained models as encoders for visual-language tasks. 
From then on, the utilization of pre-trained models in vision-language tasks became more and more popular.
For instance, Flamingo~\cite{alayrac2022flamingo} and BLIP-2~\cite{li2023blip} align the pre-trained image and text encoders employing gated cross-attention and Q-Former, respectively.
Furthermore, LLaVA~\cite{liu2023visual} and MiniGPT-4~\cite{zhu2023minigpt} leverage more robust Large Language Models (LLM)~\cite{chiang2023vicuna, touvron2023llama} as text encoders, while only training a projection layer for alignment. 
With the increasing popularity of LLM, an increasing number of studies~\cite{bai2023qwen,wang2023visionllm,zhou2023regionblip} explore the potential of multi-modal LLMs for vision-language tasks.

%% file: sec/3_method.tex
\section{Method}

\begin{figure*}
\centerline{\includegraphics[width=0.98\textwidth]{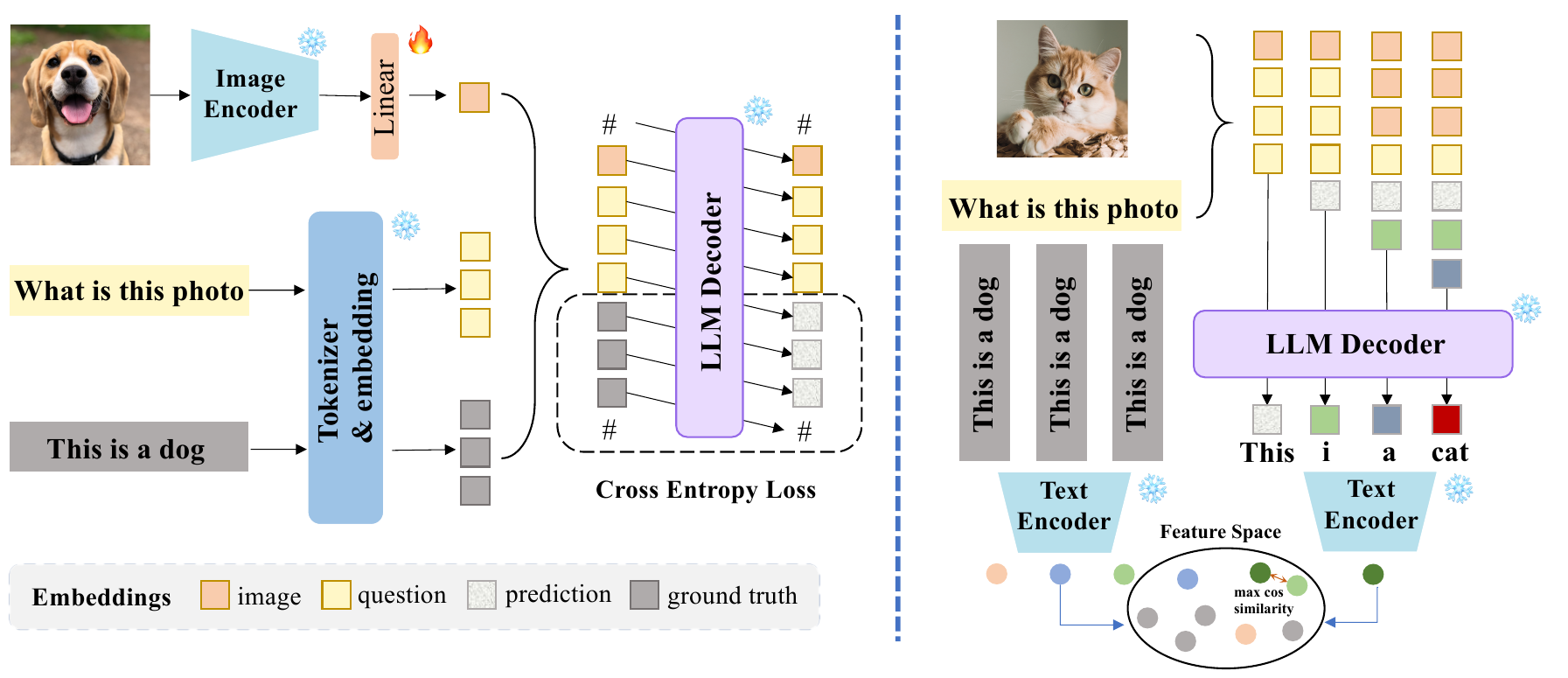}}
\caption{The overview structure of our proposed method. The conceptual illustration of the generative multi-modal model (GMM) is shown on the left. In order to adapt this model for CIL, We have to turn the GMM model for classification and further adapt to our objective benchmark for learning (see Sec.~\ref{sec:core}). On the right side, we demonstrate how the final evaluation is taken for CIL with all seen classes. 
Text encoder is used for obtaining the embeddings for similarity prediction. }
\label{fig:overview}
\end{figure*}
In this section, we introduce the preliminaries of 
class-incremental learning and generative multi-modal models. Then, we present our approach to leverage generative models for CIL and the corresponding learning process.

\subsection{Preliminaries}
\minisection{Class-Incremental Learning.}
Given N tasks $T=\{T_1, T_2, ..., T_N\}$, 
the goal of class-incremental learning is 
to learn each task $T_t$ with its associated data 
$\{\mathbf{X}_t, \mathbf{Y}_t\}$ in a sequential order. For each task, it contains samples
$\{\mathbf{x}_i, \mathbf{y}_i\}, i=1,...,n_{t}$, 
 where $\mathbf{x}_i$ is the images and $\mathbf{y}_i$ is the corresponding one-hot labels. 
Typically, $\mathbf{X}_i \cap \mathbf{X}_j = \emptyset, \forall i \neq j$. 
At inference, the model is tested on all seen tasks 
without task IDs. In some scenarios, fixed memory storage 
is set to keep a few samples of previous 
tasks to prevent forgetting.

Normally, a CIL model consists of a feature extractor 
and a classifier head $F = \{f_\theta, \mathcal{H}_\phi\}$ which 
are parameterized by $\{\theta, \phi\}$. In traditional 
class-incremental learning, $\theta$ is usually a modified 
ResNet~\cite{he2016deep} with all parameters tunable. In pre-trained or prompt-based methods, $\theta$ represents fewer trainable parameters 
like a linear adaptor or a couple of prompts. $\phi$ is a 
linear classifier head projecting image features to 
probability predictions, which has to be expanded for each new task in order to make predictions for the new classes. The conventional Cross
Entropy loss is often used for updating $\theta$ and $\phi$, which for
task t is:

\begin{equation}
    \label{eq:entropy-our}
        \mathcal{L}_{\text{CE}} (\mathbf{X}_t, \mathbf{Y}_t; \theta, \phi) = -\frac{1}{n_t} \sum_{i=1} ^{n_t} \mathbf{y}_{i}\cdot \log \mathcal{H} \left( f\left(\mathbf{x}_{i} ;\theta \right) ; \phi\right).
    \end{equation}
In the continual learning process, the parameter $\phi$ can easily deviate to the data of the current task due to the absence or scarcity of old samples in the previous tasks, resulting in forgetting the previously acquired knowledge and deteriorating the overall performance.

\minisection{Generative Multi-Modal Models (GMM).}
Multi-modal models have demonstrated exceptional performance in generating detailed image descriptions by incorporating both visual and textual information. Notably, GPT-4~\cite{openai2023gpt4} stands out as an advanced model proficient in generating comprehensive image descriptions and providing explanations for the depicted content. Furthermore, MiniGPT-4~\cite{zhu2023minigpt} proposes a two-stage fine-tuning process that 
aligns image features and large language models, enabling LLaMa~\cite{touvron2023llama} to recognize images and conduct further dialogue based on the image content.

As shown in Fig.~\ref{fig:overview}, these models consist of an encoder $f_{enc}$ to generate content embeddings including image embedding and text embedding, which is further used as the input of an auto-aggressive decoder $f_{dec}$ to generate image descriptions. Input image $\mathbf{x}_i$ is encoded as an image embedding $e_i$, and question embeddings $\mathbf{q}$ with $q_{1}, ..., q_{l}$ can be concatenated with the image embedding together to generate answer embeddings $\mathbf{s}$ with $s_{1}, ..., s_{m}$. The output tokens are generated one by one with the condition of previously generated tokens. For instance, $s_{m}$ is generated with all previous $m-1$ tokens (see the decoder in Fig.~\ref{fig:overview}).  

\subsection{Generative Multi-Modal Models for CIL}
\label{sec:core}

We adhere to the foundational settings of MiniGPT-4, incorporating a frozen image encoder $f_{enc}$ followed by a trainable projection layer for adaptation to downstream tasks, as shown in Fig.~\ref{fig:overview}. Our primary innovation involves the direct utilization of generative models to produce text, which can then serve as a basis for discriminative classification. However, two major challenges need addressing. First, there is the issue of adapting generative multi-modal models for classification, given that the generated text may differ significantly from class names. Second, we must devise a mechanism for our classification benchmarks to learn in a manner consistent with generative multi-modal models. We introduce these two aspects as follows.

\minisection{Turning GMM for classification.}
We employ a distance metric to bridge the gap between generative and discriminative models. During training, we use the real label's text to encourage the model to predict the label of an image with a concise and accurate sentence in the format of ``This is a photo of [CLS]." avoiding detailed descriptions of all contents in the image. 
During testing, the model follows the format to output the category text for a given image. We extract the content in ``[CLS]", then obtain its text features using the CLIP~\cite{clip} text encoder $f_\text{text}$, and compute the distance with text features of all categories seen by now. The closest class was then considered the final prediction of the generative model.

\minisection{Converting CIL Benchmarks for adaptation.}
CIL is usually evaluated on ImageNet, CIFAR-100, and ImageNet-R datasets. These datasets usually consist of images and corresponding one-hot labels $\{\mathbf{X}_t, \mathbf{Y}_t\}$. Using the CIFAR100 dataset as an example, 
we pair each image with a sentence to form an image-text pair 
format $\{\mathbf{X}_t, \mathbf{S}_t\}$ with the template: “This is a photo of [CLS]”, where “[CLS]” is the label name of that category, 
such as apple, dog, etc. Next, we partitioned the 100 classes into various tasks based on different settings and fed them into the model sequentially. After completing the training on task $T$, the model should be capable of classifying all the classes encompassed from task $0$ to task $T$. Note that only the linear projection layer is updated for further adaptation.

\subsection{Optimization and Inference}

\begin{table*}
    \centering
    \resizebox{0.95\textwidth}{!}{%
    \begin{tabular}{llcccccccc}
    \toprule
    \multirow{3}{*}{Type}                                                       & \multirow{3}{*}{Method}       & \multirow{3}{*}{\begin{tabular}[c]{@{}c@{}}Exemplar\end{tabular}} & \multicolumn{6}{c}{Tiny-ImageNet}                                                                                                       & ImageNet-R           \\ \cline{4-10}  
                                                                                &                             &                                                                          & \multicolumn{2}{c}{5 tasks}                 & \multicolumn{2}{c}{10 tasks}                & \multicolumn{2}{c}{20 tasks}                & 10 tasks             \\
                                                                                &                             &                                                                          & \textbf{Avg}         & \textbf{Last}        & \textbf{Avg}         & \textbf{Last}        & \textbf{Avg}         & \textbf{Last}        & Last                 \\ \hline
    \multirow{7}{*}{Conventional}                                                & EWC~\cite{kirkpatrick2017overcoming}~                        & \XSolidBrush                                                                         & 19.01                & 6.00                 & 15.82                & 3.79                 & 12.35                & 4.73                 & 35.00                 \\
                                                                                & LwF~\cite{li2017learning}          &  \XSolidBrush                                                                        & 22.31                & 7.34                 & 17.34                & 4.73                 & 12.48                & 4.26                 & 38.50                 \\
                                                                                & iCaRL~\cite{rebuffi2017icarl}  & \Checkmark                                                                         & 45.95                & 34.60                & 43.22                & 33.22                & 37.85                & 27.54                & -                    \\
                                                                                & EEIL~\cite{castro2018end}          &  \Checkmark                                                                        & 47.17                & 35.12                & 45.03                & 34.64                & 40.41                & 29.72                & -                    \\
                                                                                & UCIR~\cite{UCIR_2019_CVPR}        &     \Checkmark                                                                     & 50.30                & 39.42                & 48.58                & 37.29                & 42.84                & 30.85                & -                    \\
                                                                                & PASS~\cite{zhu2021prototype}       &   \XSolidBrush                                                                       & 49.54                & 41.64                & 47.19                & 39.27                & 42.01                & 32.93                & -                    \\
                                                                                & DyTox~\cite{douillard2022dytox} & \Checkmark                                                                         & 55.58                & 47.23                & 52.26                & 42.79                & 46.18                & 36.21                & -                    \\ \hline
    \multirow{8}{*}{\begin{tabular}[c]{@{}l@{}}Discriminative \\PT models\end{tabular}}   & Continual-CLIP\cite{continual-clip}              & \XSolidBrush                                                                         & 70.49       & 66.43      & 70.55     & 66.43       & 70.51       & 66.43       & 72.00                     \\
                                                                                & L2P~\cite{l2p}                         &  \XSolidBrush                                                                         &          83.53            &        78.32              &        76.37              &          65.78            &       68.04               &         52.40             &       72.92                \\
                                                                                & L2P~\cite{l2p}                         &  \Checkmark                                                                       &        80.24              &     72.89                 &            80.08          &        72.61              &          79.44            &       70.41               &     59.78                \\
                                                                                & DualPrompt~\cite{wang2022dualprompt}                       &  \XSolidBrush                                                                         &      85.15                &    81.01                  &          81.38            &    73.73                  &    73.45                  &      60.16                & 68.82                     \\
                                                                                & DualPrompt~\cite{wang2022dualprompt}                       &  \Checkmark                                                                        &       79.92               &   72.83                   &      79.15                & 73.21                     &    80.17                  &           71.74           &       57.02               \\
                                                                                 & CODA-Prompt~\cite{smith2023coda}                         & \XSolidBrush                                                                         &             \textbf{85.91}         &         \textbf{81.36}             &  82.80                    &   75.28                   &          77.43            &  66.32                    &   73.88                   \\
                                                                                & Linear Probe                &      \XSolidBrush                                                                     &   74.38                   &          65.40            &         69.73             &     58.31                 &          60.14            &   49.72                   &     45.17                 \\
                                                                                & Linear Probe                &   \Checkmark                                                                       &    70.10                  &          61.11            &      69.35                &   64.19                   &     71.64                 & 70.50                     &     55.72                 \\ \hline
    \multirow{3}{*}{\begin{tabular}[c]{@{}l@{}}Generative\\ PT models\end{tabular}} & Zero-shot                   &  \XSolidBrush                                                                         &       58.16               &       53.72               &    58.10                  &           53.72           &           58.13           &         53.72             &       67.38               \\
                                                                                & GMM (Ours)                        &  \XSolidBrush                                                                         & 83.42                     & 76.98                     &  82.49                    & 76.51                     & 81.70                     & 76.03                     &        80.72              \\
                                                                                & GMM (Ours)                        &  \Checkmark                                                                       & 84.16                     & 78.46                   &  \textbf{83.95}                   &  \textbf{78.64}                    &  \textbf{84.23}                    & \textbf{79.17}                     &  \textbf{89.41}                    \\
                                                                                \bottomrule
    \end{tabular}%
    }
    \caption{Comparison results of our method with other conventional baselines and methods learned with discriminative pre-trained (PT) models on Tiny-ImageNet and ImageNet-R under the conventional CIL setting. ``Avg'' represents the averaged performance after training each task, and   ``Last'' represents the performance on all test samples after training the last task.}
    \label{tab:tiny_CIL}
    \vspace{-5pt}
    \end{table*}
    
\minisection{Optimization.} 
For each task $t$, we obtain the current task’s 
image-text pair $\{\mathbf{X}_t, \mathbf{S}_t\}$, where $\mathbf{S}_t$ contains the 
corresponding sentence of each image. During training,
we first utilize a tokenizer to tokenize and acquire the embedding 
of the questions and answers.
We leverage pre-trained encoder $f_{enc}$ and the projection layer
to obtain the corresponding features for the input images:

\begin{equation}    
\mathbf{e}_i= f_{enc}(\mathbf{x}_i; \theta_{enc}).
\end{equation}

Then, the question embedding and the ground-truth embedding of this question, 
e.g., ``This is a photo of [CLS]'', is concatenated with image embedding. The final input of LLM Decoder $f_{dec}$ is:

\begin{equation}
\hat{\mathbf{e}}_i = \text{CONCATE}(bos, \mathbf{e}_i, \mathbf{q}, \mathbf{s}, eos).
\end{equation}
$bos$ is the symbol of the sentence beginning, and $eos$ is the symbol for the end of the sentence. 
This encourages tokens at positions $m-1$ to predict 
token $m$:
\begin{equation}
\small
P(\hat{s}_1, \hat{s}_2,..., \hat{s}_{m}|\mathbf{x}_i, \mathbf{q}, \mathbf{s}) = \prod \limits_{j=1}^{m-1}P(s_j|\mathbf{e}_i, \mathbf{q}, s_1, s_2,\ldots,s_{j-1}),
\end{equation}

where $s_j$ indicates the ground-truth answer token and $\hat{s}_{m}$ is the generated prediction. Then, we can compute the Cross Entropy loss as follows:

\begin{equation}
        \mathcal{L_\text{CE}} = -\frac{1}{m} \sum_{j=1} ^{m} s_{j}\cdot \log \hat{s}_{j}. 
\end{equation}

\minisection{Inference.}
During inference, we use the updated projection layer in conjunction 
with the pre-trained encoder to obtain image features. 
These image features, combined with the question 
embeddings are then passed to the LLM Decoder 
to obtain the text output.


\begin{equation}
pred = \text{argmax}<f_\text{text} (\mathbf{s}), f_\text{text} (\mathbf{\hat{s}})>, 
\end{equation}
where $f_\text{text}$ is the text encoder, $<,>$ is the cosine similarity used to calculate the final predictions $pred$.

%% file: sec/4_experiments.tex
\section{Experiments}

\subsection{Experimental setups}
\begin{figure*}[t]
    \centering
    \includegraphics[width=0.93\textwidth]
    {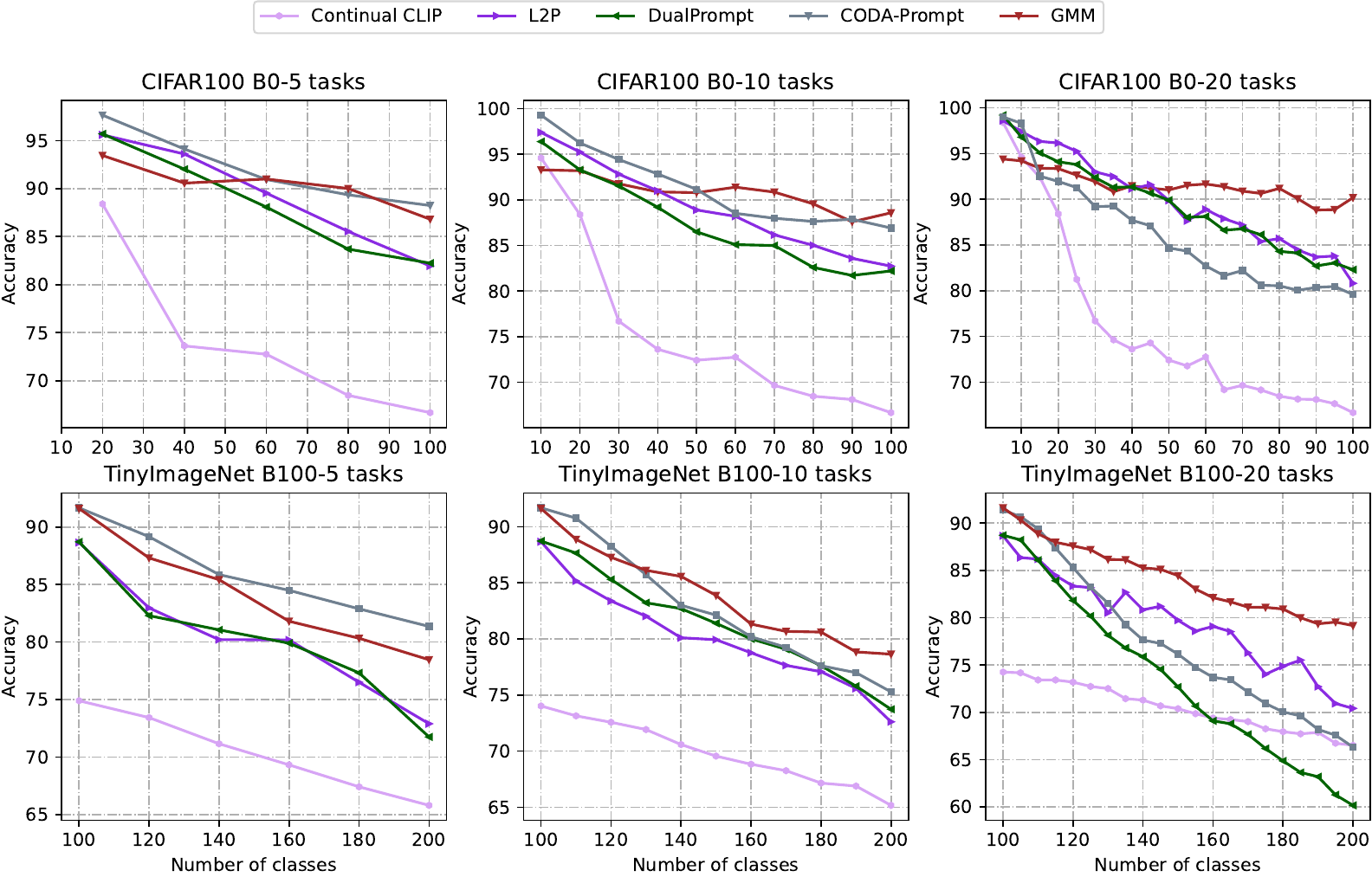}
    \vspace{-6pt}
    \caption{Comparison of our method with other 
    SOTA baselines on CIFAR100 and Tiny-ImageNet under 
    the conventional CIL setting.
    }
    \label{fig_cifar_res}
    \end{figure*}
\minisection{Datasets and Baselines.}
We conduct experiments in both conventional CIL and Few-shot CIL scenarios.
In conventional CIL, we evaluate on three datasets. 
CIFAR100, Tiny-ImageNet and ImageNet-R. 
CIFAR100 contains 60,000 images of 32x32 pixels in 100 categories. 
Each category has 600 images, of which 500 are for the 
training set, and 100 are for the test set. We experiment 
with two settings, B0-n and B50-n. The former splits 100 
classes into $n$ tasks, while the latter trains on 50 classes 
first and then distributes the other 50 classes across 5/10 tasks. 

Tiny-ImageNet contains 200 classes out of the original 1000 
classes in ImageNet, with 550 images per class, of which 500 
are in the training set and 50 are in the test set. 
The images are down-sampled to $64 \times 64$ pixels, which makes 
them easier to process and analyze. We train the first task in half 
100 classes and split the other 100 classes into 5/10/20 tasks following~\cite{zhu2021prototype}.

ImageNet-R~\cite{hendrycks2021many} contains 200 classes of images, 
which are included in the original ImageNet 1000 classes. 
However, many images are newly added and have various styles, 
such as sketch, painting, misc, etc. The dataset poses a great 
challenge for continual learning, as it has a wide diversity of
image categories and styles and an uneven distribution of samples 
ranging from 45 to 500 per category. We follow~\cite{wang2022dualprompt} 
to split the dataset into 10 tasks, each containing 20 classes.


In Few-shot CIL, we use CIFAR100 and \textit{mini}-ImageNet~\cite{russakovsky2015imagenet} following 
the split proposed by~\cite{tao2020few}. For both datasets, we partition the data into two parts: 
base session and incremental sessions. 
The base session comprises 60 classes with all data available, 
while the incremental session follows a 5-way 5-shot setting, 
meaning that each session consists of only 5 classes with 5 samples each.

In both conventional and few-shot scenarios, 
we compare our method with some of the current state-of-the-art approaches, including 
conventional methods~\cite{rebuffi2017icarl, UCIR_2019_CVPR,BiC, zhao2020maintaining,douillard2020podnet,
yan2021dynamically, douillard2022dytox,shi2021overcoming, li2017learning, castro2018end, zhu2021prototype}, 
pre-trained and prompt-based methods~\cite{l2p,wang2022dualprompt,smith2023coda}, 
and some specifically designed methods~\cite{tao2020few, zhang2021few, shi2021overcoming, chi2022metafscil} 
for few-shot scenarios. Additionally, we compare with a linear probe baseline, 
wherein features obtained from the image encoder are connected to a classifier for 
classification. We also consider the Zero-shot approach, where the generated text is 
directly used for classification without further fine-tuning.

\minisection{Implementation Details.}
We follow BLIP2~\cite{li2023blip} to use the EVA-CLIP~\cite{sun2023eva} pre-trained ViT-g/14 and BLIP2 pre-trained Qfomer. 
We also used the MiniGPT-4 pre-trained projection layer checkpoint as our initial parameters.
Under the many-shot ``B0'' setting, 
we employ a learning rate of 3e-7 and use a scheduler with cosine decay. 
The total training process consists of 2 epochs only.
In B50 or B100 settings, we first train the linear layer 
with a learning rate of 3e-6 on the base classes, 
and then on the subsequent tasks, we adopt a lower learning rate of 3e-7, both employing a cosine decay scheduler.
For the few-shot setting, we use a learning rate of 3e-6 for both 
base task and incremental tasks. We train one epoch for the base task and 
two epochs for incremental tasks.

\begin{table*}
\centering

\resizebox{0.9\textwidth}{!}{\begin{tblr}{
  hline{1} = {-}{0.08em},
  hline{2, 13} = {-}{},
  hline{15} = {-}{0.08em},
}
            & 0 & 1     & 2     & 3     & 4     & 5     & 6     & 7     & 8     & PD↓   \\
iCaRL~\cite{rebuffi2017icarl}       & 61.31   & 46.32 & 42.94 & 37.63 & 30.49 & 24.00    & 20.89 & 18.80  & 17.21 & 44.10  \\
EEIL~\cite{castro2018end}        & 61.31   & 46.58 & 44.00    & 37.29 & 33.14 & 27.12 & 24.10  & 21.57 & 19.58 & 41.73 \\
LUCIR~\cite{UCIR_2019_CVPR}       & 61.31   & 47.80  & 39.31 & 31.91 & 25.68 & 21.35 & 18.67 & 17.24 & 14.17 & 47.14 \\
TOPIC~\cite{tao2020few}       & 61.31   & 50.09 & 45.17 & 41.16 & 37.48 & 35.52 & 32.19 & 29.46 & 24.42 & 36.89 \\
CEC~\cite{zhang2021few}         & 72.00      & 66.83 & 62.97 & 59.43 & 56.70  & 53.73 & 51.19 & 49.24 & 47.63 & 24.37 \\
F2M~\cite{shi2021overcoming}         & 72.05   & 67.47 & 63.16 & 59.70  & 56.71 & 53.77 & 51.11 & 49.21 & 47.84 & 24.21 \\
MetaFSCIL~\cite{chi2022metafscil}   & 72.04   & 67.94 & 63.77 & 60.29 & 57.58 & 55.16 & 52.90  & 50.79 & 49.19 & 22.85 \\
Entropy-reg~\cite{liu2022few} & 71.84   & 67.12 & 63.21 & 59.77 & 57.01 & 53.95 & 51.55 & 49.52 & 48.21 & 23.63 \\
L2P$^{*}$~\cite{l2p} & 94.12   &87.20 & 80.99 & 75.67 & 70.94 & 66.76 & 63.11 & 59.81 & 56.83 &  37.29\\
DualPrompt$^{*}$~\cite{wang2022dualprompt} & 93.97   & 86.85 & 80.67 & 75.31 & 70.61 & 66.44 & 62.77 & 59.58 & 56.80 & 37.17 \\
CODA-Prompt$^{*}$~\cite{smith2023coda} & \textbf{95.37}   & \textbf{88.86} & 82.69 & 77.87 & 74.47 & 70.16 & 66.46 & 63.73 & 61.14 & 34.23 \\
Zero-shot & 58.08   & 58.95 & 57.76 & 57.89 & 58.19 & 57.42 & 56.26 & 54.82 & 54.95 & \textbf{3.13} \\
GMM (Ours)        &  89.35  & 88.40 & \textbf{86.11} &\textbf{85.07}  &\textbf{83.61} & \textbf{81.35} & \textbf{78.97} & \textbf{77.34} & \textbf{75.18} & 14.17
\end{tblr}}
\caption{Comparison results of our method with other SOTA baselines 
on \textit{mini}-ImageNet under the 
few-shot CIL setting. Phase 0 is the base task with 
all samples available, and the following 1-8 phases are 
the incremental 5-way 5-shot tasks. The reported accuracy 
is the test result on all seen classes after  
each session of training. ``PD'' represents Performance 
Prop between session 0 and session 8, 
with lower values indicating lower forgetting. $^{*}$ indicates our re-implementation based on PILOT~\cite{sun2023pilot}.}
\label{tab:mini_few}
\end{table*}

\subsection{Experiments on Conventional CIL}
\begin{figure*}
\centering
\includegraphics[width=0.950\textwidth]
{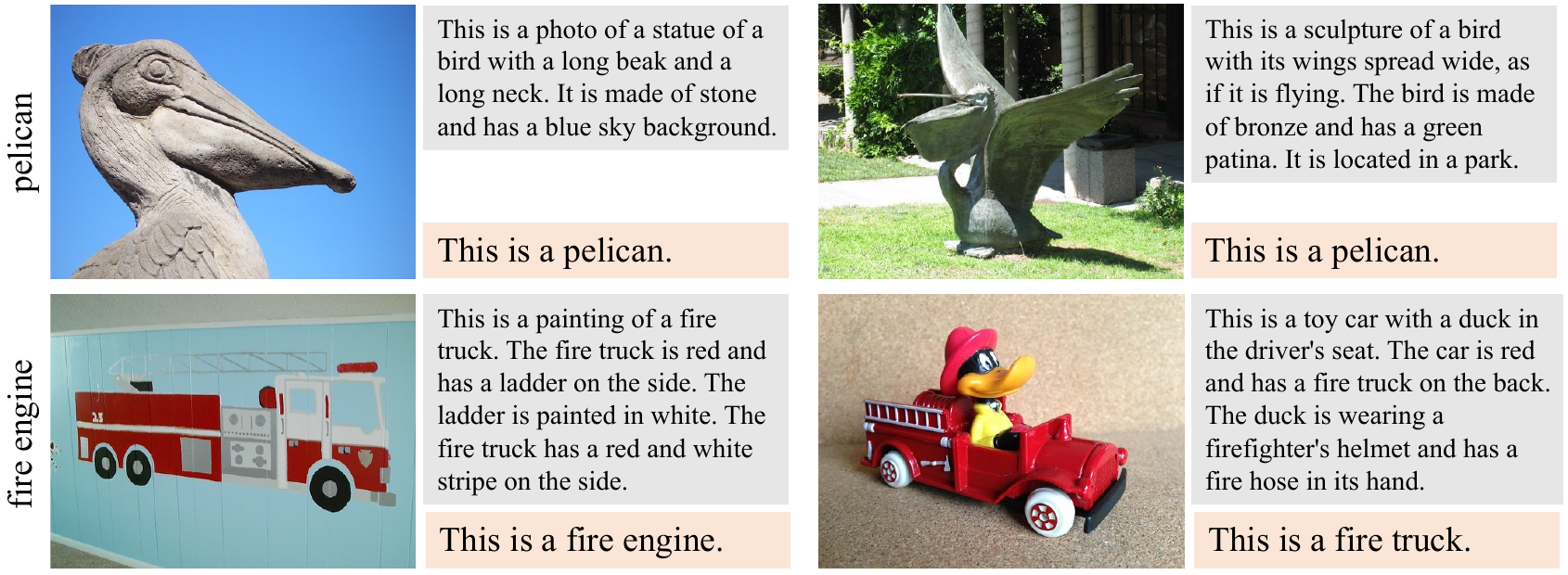}
\caption{Visual comparison examples of our method against frozen MiniGPT-4~\cite{zhu2023minigpt} (Zero-shot).  Text with a gray background is generated by MiniGPT-4 based on the image, while text with a orange background represents our method's output. The ground-truth labels are displayed on the left side of each row of images. All images displayed here are random sampled from ImageNet-R.
}
\vspace{-9pt
}
\label{fig_vis}
\end{figure*}

In Table~\ref{tab:tiny_CIL}, we can see that 
our method outperforms all conventional methods by a 
large margin, including ResNet-based method DER and ViT-based DyTox.  Note that without exemplar, our performance is a bit lower than DualPrompt and CODA-Prompt 
at B100-5 setting. 
We argue that their performance is mainly due to the backbone pre-trained on ImageNet-21K, 
which largely overlaps with CIFAR100 and Tiny-ImageNet. Another interesting observation is that our method has better performance than
all baselines under longer sequence settings (B100-10, B100-20). We believe  
this is because generative models do not rely on classification heads, 
making them less prone to bias toward the current task, 
resulting in less forgetting of past tasks. The Linear probe setting performs less than our method, indicating that our 
main contribution is not from the Large pre-trained ViT but the generation pipeline. 
In addition, the Zero-shot performance is superior to many traditional baselines, 
meaning that the Generative Multi-modal Models are indeed efficient Class-Incremental 
learners, but its output is less concise without fine-tuning (see Fig.~\ref{fig_vis}).

In Fig.~\ref{fig_cifar_res}, we compared our approach with some 
pre-trained models on CIFAR100 and Tiny-ImageNet in terms of 
last task accuracy (all baselines are based on PILOT~\cite{sun2023pilot} and use 2000 exemplars).
It can be observed that our method does not outperform other approaches 
in the initial tasks (0-2) and short sequence settings (B0-5, B100-5). 
This is because we do not rely on a supervised ImageNet-21K pre-trained backbone. 
Besides, 
we trained each task for only 1-2 epochs to ensure efficiency without sacrificing generalization. 
However, our method exhibits significant advantages in long sequences and later tasks. 
For instance, under the CIFAR100 B0-20 setting, 
we outperform CODA-Prompt by 10 points and DualPrompt by 7 points.

\begin{table}
    \centering
    \resizebox{0.46\textwidth}{!}{\begin{tblr}{
      hline{1} = {-}{0.08em},
  hline{2, 13} = {-}{},
  hline{15} = {-}{0.08em},
    }
    Method      & 0 & 4     & 8     & PD↓   \\
    iCaRL~\cite{rebuffi2017icarl}       & 64.10     & 27.93 & 13.73 & 50.37 \\
    EEIL~\cite{castro2018end}        & 64.10     & 28.96 & 15.85 & 48.25 \\
    LUCIR~\cite{UCIR_2019_CVPR}       & 64.10     & 31.61 & 13.54 & 50.56 \\
    TOPIC~\cite{tao2020few}       & 64.10     & 40.11 & 29.37 & 34.73 \\
    CEC~\cite{zhang2021few}         & 73.07    & 58.09 & 49.14 & 23.93 \\
    F2M~\cite{shi2021overcoming}         & 71.45    & 57.76 & 49.35 & 22.06 \\
    MetaFSCIL~\cite{chi2022metafscil}   & 74.50     & 59.48 & 49.97 & 24.53 \\
    Entropy-reg~\cite{liu2022few} & 74.40     & 59.71 & 50.14 & 24.26 \\
    L2P$^{*}$~\cite{l2p}         & 91.22    & 68.66 & 54.89 & 36.33 \\
    DualPrompt$^{*}$~\cite{wang2022dualprompt}  & 91.08    & 68.45 & 54.67 & 36.41 \\
    CODA-Prompt$^{*}$~\cite{smith2023coda} & \textbf{93.55}    & 71.91 & 59.32 & 34.23 \\
    Zero-shot & 74.13    & 72.59 & 67.93 & \textbf{6.20} \\
    GMM (Ours)        & 91.53    & \textbf{85.65} & \textbf{81.47} & 10.06 
    \end{tblr}}
    \caption{Comparison results of our method with other SOTA baselines on CIFAR100 under the few-shot CIL setting.}
    \vspace{-15pt}
    \label{tab:cifar_few}
    \end{table}

\subsection{Experiments on Few-shot CIL }
In Table~\ref{tab:mini_few}, we compare our method with several baselines in the few-shot setting on \textit{mini}-ImageNet. The evaluation metric is the model's accuracy across all the classes it has encountered so far.
 Our method outperforms conventional 
 methods by a substantial margin in the final task, 
 achieving an increase of more than 26\%. Furthermore, we surpass the best 
 discriminative pre-trained approach CODA-Prompt by more then 14\% points. 
 It's important to note that our accuracy in the 
 first task (89.35) might not be as high as CODA-Prompt (95.37). 
 However, in subsequent sessions, we consistently perform better than CODA-Prompt due to our ability 
 to learn new tasks and retain knowledge 
 of old tasks simultaneously.

In Table~\ref{tab:cifar_few}, our method outperforms all other baselines on the CIFAR100 dataset of the few-shot setting, achieving a remarkably lower Performance Drop (PD) of 10.06.
Furthermore, the Zero-shot baseline can achieve a very low PD due to the absence of forgetting. However, its overall performance is not very satisfying, as the length and content of its output are inconsistent and unpredictable without fine-tuning.



\subsection{Visualizations}
In Fig.~\ref{fig_vis}, we present some comparison examples of our methods against GMM without fine-tuning~\cite{zhu2023minigpt}. We can see that GMM without fine-tuning provides an intuitive description of the overall image content with varying lengths of output text. However, it tends to recognize only broad categories (e.g., biar, car) and struggles with fine-grained categorization (e.g., pelican, truck). The descriptions are sometimes somewhat repetitive (e.g., first fire engine). In contrast, our fine-tuned method accurately identifies the image's real category, even if there are occasional discrepancies with the true labels (e.g., "fire engine" vs. "fire truck"). Besides, with the assistance of the text encoder during the testing phase, our model can achieve correct classification results even when predicting similar but not identical text.

%% file: arxiv.bbl
\begin{thebibliography}{90}
\providecommand{\natexlab}[1]{#1}
\providecommand{\url}[1]{\texttt{#1}}
\expandafter\ifx\csname urlstyle\endcsname\relax
  \providecommand{\doi}[1]{doi: #1}\else
  \providecommand{\doi}{doi: \begingroup \urlstyle{rm}\Url}\fi

\bibitem[Ahn et~al.(2021)Ahn, Kwak, Lim, Bang, Kim, and Moon]{ahn2021ss}
Hongjoon Ahn, Jihwan Kwak, Subin Lim, Hyeonsu Bang, Hyojun Kim, and Taesup Moon.
\newblock Ss-il: Separated softmax for incremental learning.
\newblock In \emph{ICCV}, 2021.

\bibitem[Alayrac et~al.(2022)Alayrac, Donahue, Luc, Miech, Barr, Hasson, Lenc, Mensch, Millican, Reynolds, et~al.]{alayrac2022flamingo}
Jean-Baptiste Alayrac, Jeff Donahue, Pauline Luc, Antoine Miech, Iain Barr, Yana Hasson, Karel Lenc, Arthur Mensch, Katherine Millican, Malcolm Reynolds, et~al.
\newblock Flamingo: a visual language model for few-shot learning.
\newblock \emph{Advances in Neural Information Processing Systems}, 35:\penalty0 23716--23736, 2022.

\bibitem[Aljundi et~al.(2017)Aljundi, Chakravarty, and Tuytelaars]{aljundi2017expert}
Rahaf Aljundi, Punarjay Chakravarty, and Tinne Tuytelaars.
\newblock Expert gate: Lifelong learning with a network of experts.
\newblock In \emph{Proceedings of the IEEE conference on computer vision and pattern recognition}, pages 3366--3375, 2017.

\bibitem[Bai et~al.(2023)Bai, Bai, Yang, Wang, Tan, Wang, Lin, Zhou, and Zhou]{bai2023qwen}
Jinze Bai, Shuai Bai, Shusheng Yang, Shijie Wang, Sinan Tan, Peng Wang, Junyang Lin, Chang Zhou, and Jingren Zhou.
\newblock Qwen-vl: A versatile vision-language model for understanding, localization, text reading, and beyond.
\newblock \emph{arXiv preprint arXiv:2308.12966}, 2\penalty0 (3):\penalty0 4, 2023.

\bibitem[Ben-Younes et~al.(2017)Ben-Younes, Cadene, Cord, and Thome]{ben2017mutan}
Hedi Ben-Younes, R{\'e}mi Cadene, Matthieu Cord, and Nicolas Thome.
\newblock Mutan: Multimodal tucker fusion for visual question answering.
\newblock In \emph{Proceedings of the IEEE international conference on computer vision}, pages 2612--2620, 2017.

\bibitem[Caron et~al.(2021)Caron, Touvron, Misra, J{\'e}gou, Mairal, Bojanowski, and Joulin]{caron2021emerging}
Mathilde Caron, Hugo Touvron, Ishan Misra, Herv{\'e} J{\'e}gou, Julien Mairal, Piotr Bojanowski, and Armand Joulin.
\newblock Emerging properties in self-supervised vision transformers.
\newblock In \emph{Proceedings of the IEEE/CVF international conference on computer vision}, pages 9650--9660, 2021.

\bibitem[Castro et~al.(2018)Castro, Mar{\'\i}n-Jim{\'e}nez, Guil, Schmid, and Alahari]{castro2018end}
Francisco~M Castro, Manuel~J Mar{\'\i}n-Jim{\'e}nez, Nicol{\'a}s Guil, Cordelia Schmid, and Karteek Alahari.
\newblock End-to-end incremental learning.
\newblock In \emph{Proceedings of the European conference on computer vision (ECCV)}, pages 233--248, 2018.

\bibitem[Chen et~al.(2022)Chen, Guo, Yi, Li, and Elhoseiny]{chen2022visualgpt}
Jun Chen, Han Guo, Kai Yi, Boyang Li, and Mohamed Elhoseiny.
\newblock Visualgpt: Data-efficient adaptation of pretrained language models for image captioning.
\newblock In \emph{Proceedings of the IEEE/CVF Conference on Computer Vision and Pattern Recognition}, pages 18030--18040, 2022.

\bibitem[Chen et~al.(2020)Chen, Fan, Girshick, and He]{chen2020improved}
Xinlei Chen, Haoqi Fan, Ross Girshick, and Kaiming He.
\newblock Improved baselines with momentum contrastive learning.
\newblock \emph{arXiv preprint arXiv:2003.04297}, 2020.

\bibitem[Chen* et~al.(2021)Chen*, Xie*, and He]{chen2021mocov3}
Xinlei Chen*, Saining Xie*, and Kaiming He.
\newblock An empirical study of training self-supervised vision transformers.
\newblock \emph{arXiv preprint arXiv:2104.02057}, 2021.

\bibitem[Cheraghian et~al.(2021)Cheraghian, Rahman, Fang, Roy, Petersson, and Harandi]{cheraghian2021semantic}
Ali Cheraghian, Shafin Rahman, Pengfei Fang, Soumava~Kumar Roy, Lars Petersson, and Mehrtash Harandi.
\newblock Semantic-aware knowledge distillation for few-shot class-incremental learning.
\newblock In \emph{Proceedings of the IEEE/CVF conference on computer vision and pattern recognition}, pages 2534--2543, 2021.

\bibitem[Chi et~al.(2022)Chi, Gu, Liu, Wang, Yu, and Tang]{chi2022metafscil}
Zhixiang Chi, Li Gu, Huan Liu, Yang Wang, Yuanhao Yu, and Jin Tang.
\newblock Metafscil: A meta-learning approach for few-shot class incremental learning.
\newblock In \emph{Proceedings of the IEEE/CVF conference on computer vision and pattern recognition}, pages 14166--14175, 2022.

\bibitem[Chiang et~al.(2023)Chiang, Li, Lin, Sheng, Wu, Zhang, Zheng, Zhuang, Zhuang, Gonzalez, et~al.]{chiang2023vicuna}
Wei-Lin Chiang, Zhuohan Li, Zi Lin, Ying Sheng, Zhanghao Wu, Hao Zhang, Lianmin Zheng, Siyuan Zhuang, Yonghao Zhuang, Joseph~E Gonzalez, et~al.
\newblock Vicuna: An open-source chatbot impressing gpt-4 with 90\%* chatgpt quality.
\newblock \emph{See https://vicuna. lmsys. org (accessed 14 April 2023)}, 2023.

\bibitem[Delange et~al.(2021)Delange, Aljundi, Masana, Parisot, Jia, Leonardis, Slabaugh, and Tuytelaars]{delange2021continual}
Matthias Delange, Rahaf Aljundi, Marc Masana, Sarah Parisot, Xu Jia, Ales Leonardis, Greg Slabaugh, and Tinne Tuytelaars.
\newblock A continual learning survey: Defying forgetting in classification tasks.
\newblock \emph{TPAMI}, 2021.

\bibitem[Douillard et~al.(2020)Douillard, Cord, Ollion, Robert, and Valle]{douillard2020podnet}
Arthur Douillard, Matthieu Cord, Charles Ollion, Thomas Robert, and Eduardo Valle.
\newblock Podnet: Pooled outputs distillation for small-tasks incremental learning.
\newblock In \emph{ECCV}, 2020.

\bibitem[Douillard et~al.(2022)Douillard, Ram{\'e}, Couairon, and Cord]{douillard2022dytox}
Arthur Douillard, Alexandre Ram{\'e}, Guillaume Couairon, and Matthieu Cord.
\newblock Dytox: Transformers for continual learning with dynamic token expansion.
\newblock In \emph{CVPR}, 2022.

\bibitem[Fernando et~al.(2017)Fernando, Banarse, Blundell, Zwols, Ha, Rusu, Pritzel, and Wierstra]{fernando2017pathnet}
Chrisantha Fernando, Dylan Banarse, Charles Blundell, Yori Zwols, David Ha, Andrei~A Rusu, Alexander Pritzel, and Daan Wierstra.
\newblock Pathnet: Evolution channels gradient descent in super neural networks.
\newblock \emph{arXiv preprint arXiv:1701.08734}, 2017.

\bibitem[Gao and Liu(2023)]{gao2023ddgr}
Rui Gao and Weiwei Liu.
\newblock Ddgr: Continual learning with deep diffusion-based generative replay.
\newblock In \emph{ICML}, 2023.

\bibitem[Guo et~al.(2023)Guo, Lu, Liu, Cheng, and Hu]{guo2023visual}
Meng-Hao Guo, Cheng-Ze Lu, Zheng-Ning Liu, Ming-Ming Cheng, and Shi-Min Hu.
\newblock Visual attention network.
\newblock \emph{Computational Visual Media}, 9\penalty0 (4):\penalty0 733--752, 2023.

\bibitem[Hayes et~al.(2020)Hayes, Kafle, Shrestha, Acharya, and Kanan]{hayes2020remind}
Tyler~L Hayes, Kushal Kafle, Robik Shrestha, Manoj Acharya, and Christopher Kanan.
\newblock Remind your neural network to prevent catastrophic forgetting.
\newblock In \emph{ECCV}, 2020.

\bibitem[He et~al.(2016)He, Zhang, Ren, and Sun]{he2016deep}
Kaiming He, Xiangyu Zhang, Shaoqing Ren, and Jian Sun.
\newblock Deep residual learning for image recognition.
\newblock In \emph{Proceedings of the IEEE conference on computer vision and pattern recognition}, pages 770--778, 2016.

\bibitem[He et~al.(2020)He, Fan, Wu, Xie, and Girshick]{he2020momentum}
Kaiming He, Haoqi Fan, Yuxin Wu, Saining Xie, and Ross Girshick.
\newblock Momentum contrast for unsupervised visual representation learning.
\newblock In \emph{Proceedings of the IEEE/CVF conference on computer vision and pattern recognition}, pages 9729--9738, 2020.

\bibitem[Hendrycks et~al.(2021)Hendrycks, Basart, Mu, Kadavath, Wang, Dorundo, Desai, Zhu, Parajuli, Guo, Song, Steinhardt, and Gilmer]{hendrycks2021many}
Dan Hendrycks, Steven Basart, Norman Mu, Saurav Kadavath, Frank Wang, Evan Dorundo, Rahul Desai, Tyler Zhu, Samyak Parajuli, Mike Guo, Dawn Song, Jacob Steinhardt, and Justin Gilmer.
\newblock The many faces of robustness: A critical analysis of out-of-distribution generalization.
\newblock \emph{ICCV}, 2021.

\bibitem[Hou et~al.(2019)Hou, Pan, Loy, Wang, and Lin]{UCIR_2019_CVPR}
Saihui Hou, Xinyu Pan, Chen~Change Loy, Zilei Wang, and Dahua Lin.
\newblock Learning a unified classifier incrementally via rebalancing.
\newblock In \emph{CVPR}, 2019.

\bibitem[Hu et~al.(2021)Hu, Tang, Miao, Hua, and Zhang]{hu2021distilling}
Xinting Hu, Kaihua Tang, Chunyan Miao, Xian-Sheng Hua, and Hanwang Zhang.
\newblock Distilling causal effect of data in class-incremental learning.
\newblock In \emph{CVPR}, 2021.

\bibitem[Kirkpatrick et~al.(2017)Kirkpatrick, Pascanu, Rabinowitz, Veness, Desjardins, Rusu, Milan, Quan, Ramalho, Grabska-Barwinska, et~al.]{kirkpatrick2017overcoming}
James Kirkpatrick, Razvan Pascanu, Neil Rabinowitz, Joel Veness, Guillaume Desjardins, Andrei~A Rusu, Kieran Milan, John Quan, Tiago Ramalho, Agnieszka Grabska-Barwinska, et~al.
\newblock Overcoming catastrophic forgetting in neural networks.
\newblock \emph{Proceedings of the national academy of sciences}, 114\penalty0 (13):\penalty0 3521--3526, 2017.

\bibitem[Li et~al.(2023)Li, Li, Savarese, and Hoi]{li2023blip}
Junnan Li, Dongxu Li, Silvio Savarese, and Steven Hoi.
\newblock Blip-2: Bootstrapping language-image pre-training with frozen image encoders and large language models.
\newblock \emph{arXiv preprint arXiv:2301.12597}, 2023.

\bibitem[Li et~al.(2019)Li, Zhang, Zhang, Huang, He, Lyu, and Gao]{li2019object}
Wenbo Li, Pengchuan Zhang, Lei Zhang, Qiuyuan Huang, Xiaodong He, Siwei Lyu, and Jianfeng Gao.
\newblock Object-driven text-to-image synthesis via adversarial training.
\newblock In \emph{Proceedings of the IEEE/CVF Conference on Computer Vision and Pattern Recognition}, pages 12174--12182, 2019.

\bibitem[Li and Hoiem(2017)]{li2017learning}
Zhizhong Li and Derek Hoiem.
\newblock Learning without forgetting.
\newblock \emph{IEEE transactions on pattern analysis and machine intelligence}, 40\penalty0 (12):\penalty0 2935--2947, 2017.

\bibitem[Li and Hoiem(2018)]{li2018learning}
Zhizhong Li and Derek Hoiem.
\newblock Learning without forgetting.
\newblock \emph{TPAMI}, 2018.

\bibitem[Liu et~al.(2022{\natexlab{a}})Liu, Gu, Chi, Wang, Yu, Chen, and Tang]{liu2022few}
Huan Liu, Li Gu, Zhixiang Chi, Yang Wang, Yuanhao Yu, Jun Chen, and Jin Tang.
\newblock Few-shot class-incremental learning via entropy-regularized data-free replay.
\newblock In \emph{European Conference on Computer Vision}, pages 146--162. Springer, 2022{\natexlab{a}}.

\bibitem[Liu et~al.(2023{\natexlab{a}})Liu, Li, Wu, and Lee]{liu2023visual}
Haotian Liu, Chunyuan Li, Qingyang Wu, and Yong~Jae Lee.
\newblock Visual instruction tuning.
\newblock In \emph{NeurIPS}, 2023{\natexlab{a}}.

\bibitem[Liu et~al.(2023{\natexlab{b}})Liu, Wei, Wu, Zuo, and Zhang]{liu2023survey}
Ming Liu, Yuxiang Wei, Xiaohe Wu, Wangmeng Zuo, and Lei Zhang.
\newblock Survey on leveraging pre-trained generative adversarial networks for image editing and restoration.
\newblock \emph{Science China Information Sciences}, 66\penalty0 (5):\penalty0 151101, 2023{\natexlab{b}}.

\bibitem[Liu et~al.(2021{\natexlab{a}})Liu, Chen, Guo, Zhu, and Liu]{liu2021cptr}
Wei Liu, Sihan Chen, Longteng Guo, Xinxin Zhu, and Jing Liu.
\newblock Cptr: Full transformer network for image captioning.
\newblock \emph{arXiv preprint arXiv:2101.10804}, 2021{\natexlab{a}}.

\bibitem[Liu et~al.(2018)Liu, Masana, Herranz, Van~de Weijer, Lopez, and Bagdanov]{liu2018rotate}
Xialei Liu, Marc Masana, Luis Herranz, Joost Van~de Weijer, Antonio~M Lopez, and Andrew~D Bagdanov.
\newblock Rotate your networks: Better weight consolidation and less catastrophic forgetting.
\newblock In \emph{ICPR}, 2018.

\bibitem[Liu et~al.(2022{\natexlab{b}})Liu, Hu, Cao, Bagdanov, Li, and Cheng]{liu2022long}
Xialei Liu, Yu-Song Hu, Xu-Sheng Cao, Andrew~D Bagdanov, Ke Li, and Ming-Ming Cheng.
\newblock Long-tailed class incremental learning.
\newblock In \emph{European Conference on Computer Vision}, pages 495--512. Springer, 2022{\natexlab{b}}.

\bibitem[Liu et~al.(2023{\natexlab{c}})Liu, Cao, Lu, Xiao, Bagdanov, and Cheng]{liu2023class}
Xialei Liu, Xusheng Cao, Haori Lu, Jia-wen Xiao, Andrew~D Bagdanov, and Ming-Ming Cheng.
\newblock Class incremental learning with pre-trained vision-language models.
\newblock \emph{arXiv preprint arXiv:2310.20348}, 2023{\natexlab{c}}.

\bibitem[Liu et~al.(2021{\natexlab{b}})Liu, Schiele, and Sun]{liu2021adaptive}
Yaoyao Liu, Bernt Schiele, and Qianru Sun.
\newblock Adaptive aggregation networks for class-incremental learning.
\newblock In \emph{CVPR}, 2021{\natexlab{b}}.

\bibitem[Lu et~al.(2019)Lu, Batra, Parikh, and Lee]{lu2019vilbert}
Jiasen Lu, Dhruv Batra, Devi Parikh, and Stefan Lee.
\newblock Vilbert: Pretraining task-agnostic visiolinguistic representations for vision-and-language tasks.
\newblock \emph{Advances in neural information processing systems}, 32, 2019.

\bibitem[Mallya et~al.(2018)Mallya, Davis, and Lazebnik]{mallya2018piggyback}
Arun Mallya, Dillon Davis, and Svetlana Lazebnik.
\newblock Piggyback: Adapting a single network to multiple tasks by learning to mask weights.
\newblock In \emph{ECCV}, 2018.

\bibitem[Masana et~al.(2022)Masana, Liu, Twardowski, Menta, Bagdanov, and Van De~Weijer]{masana2022class}
Marc Masana, Xialei Liu, Bart{\l}omiej Twardowski, Mikel Menta, Andrew~D Bagdanov, and Joost Van De~Weijer.
\newblock Class-incremental learning: survey and performance evaluation on image classification.
\newblock \emph{IEEE Transactions on Pattern Analysis and Machine Intelligence}, 45\penalty0 (5):\penalty0 5513--5533, 2022.

\bibitem[Mazumder et~al.(2021)Mazumder, Singh, and Rai]{mazumder2021few}
Pratik Mazumder, Pravendra Singh, and Piyush Rai.
\newblock Few-shot lifelong learning.
\newblock In \emph{Proceedings of the AAAI Conference on Artificial Intelligence}, pages 2337--2345, 2021.

\bibitem[McCloskey and Cohen(1989)]{mccloskey1989catastrophic}
Michael McCloskey and Neal~J Cohen.
\newblock Catastrophic interference in connectionist networks: The sequential learning problem.
\newblock In \emph{Psychology of learning and motivation}, pages 109--165. Elsevier, 1989.

\bibitem[OpenAI(2023)]{openai2023gpt4}
OpenAI.
\newblock Gpt-4 technical report, 2023.

\bibitem[Qu et~al.(2021)Qu, Rahmani, Xu, Williams, and Liu]{qu2021recent}
Haoxuan Qu, Hossein Rahmani, Li Xu, Bryan Williams, and Jun Liu.
\newblock Recent advances of continual learning in computer vision: An overview.
\newblock \emph{arXiv preprint arXiv:2109.11369}, 2021.

\bibitem[Radford et~al.(2021)Radford, Kim, Hallacy, Ramesh, Goh, Agarwal, Sastry, Askell, Mishkin, Clark, et~al.]{clip}
Alec Radford, Jong~Wook Kim, Chris Hallacy, Aditya Ramesh, Gabriel Goh, Sandhini Agarwal, Girish Sastry, Amanda Askell, Pamela Mishkin, Jack Clark, et~al.
\newblock Learning transferable visual models from natural language supervision.
\newblock In \emph{ICML}, 2021.

\bibitem[Rajasegaran et~al.(2019)Rajasegaran, Hayat, Khan, Khan, and Shao]{rajasegaran2019random}
Jathushan Rajasegaran, Munawar Hayat, Salman Khan, Fahad~Shahbaz Khan, and Ling Shao.
\newblock Random path selection for incremental learning.
\newblock In \emph{NIPS}, 2019.

\bibitem[Rebuffi et~al.(2017)Rebuffi, Kolesnikov, Sperl, and Lampert]{rebuffi2017icarl}
Sylvestre-Alvise Rebuffi, Alexander Kolesnikov, Georg Sperl, and Christoph~H Lampert.
\newblock icarl: Incremental classifier and representation learning.
\newblock In \emph{CVPR}, 2017.

\bibitem[Russakovsky et~al.(2015)Russakovsky, Deng, Su, Krause, Satheesh, Ma, Huang, Karpathy, Khosla, Bernstein, et~al.]{russakovsky2015imagenet}
Olga Russakovsky, Jia Deng, Hao Su, Jonathan Krause, Sanjeev Satheesh, Sean Ma, Zhiheng Huang, Andrej Karpathy, Aditya Khosla, Michael Bernstein, et~al.
\newblock Imagenet large scale visual recognition challenge.
\newblock \emph{International journal of computer vision}, 115:\penalty0 211--252, 2015.

\bibitem[Schwarz et~al.(2018)Schwarz, Czarnecki, Luketina, Grabska-Barwinska, Whye~Teh, Pascanu, and Hadsell]{schwarz2018progress}
Jonathan Schwarz, Wojciech Czarnecki, Jelena Luketina, Agnieszka Grabska-Barwinska, Yee Whye~Teh, Razvan Pascanu, and Raia Hadsell.
\newblock Progress \& compress: A scalable framework for continual learning.
\newblock In \emph{ICML}, 2018.

\bibitem[Serra et~al.(2018)Serra, Suris, Miron, and Karatzoglou]{serra2018overcoming}
Joan Serra, Didac Suris, Marius Miron, and Alexandros Karatzoglou.
\newblock Overcoming catastrophic forgetting with hard attention to the task.
\newblock In \emph{ICML}, 2018.

\bibitem[Shao et~al.(2023)Shao, Guo, Zhao, and Liu]{shao2023class}
Yijia Shao, Yiduo Guo, Dongyan Zhao, and Bing Liu.
\newblock Class-incremental learning based on label generation.
\newblock In \emph{Proceedings of the 61st Annual Meeting of the Association for Computational Linguistics (Volume 2: Short Papers)}, pages 1263--1276, Toronto, Canada, 2023. Association for Computational Linguistics.

\bibitem[Shi et~al.(2021)Shi, Chen, Zhang, Zhan, and Wu]{shi2021overcoming}
Guangyuan Shi, Jiaxin Chen, Wenlong Zhang, Li-Ming Zhan, and Xiao-Ming Wu.
\newblock Overcoming catastrophic forgetting in incremental few-shot learning by finding flat minima.
\newblock \emph{Advances in neural information processing systems}, 34:\penalty0 6747--6761, 2021.

\bibitem[Shin et~al.(2017)Shin, Lee, Kim, and Kim]{shin2017continual}
Hanul Shin, Jung~Kwon Lee, Jaehong Kim, and Jiwon Kim.
\newblock Continual learning with deep generative replay.
\newblock \emph{Advances in neural information processing systems}, 30, 2017.

\bibitem[Smith et~al.(2023)Smith, Karlinsky, Gutta, Cascante-Bonilla, Kim, Arbelle, Panda, Feris, and Kira]{smith2023coda}
James~Seale Smith, Leonid Karlinsky, Vyshnavi Gutta, Paola Cascante-Bonilla, Donghyun Kim, Assaf Arbelle, Rameswar Panda, Rogerio Feris, and Zsolt Kira.
\newblock Coda-prompt: Continual decomposed attention-based prompting for rehearsal-free continual learning.
\newblock In \emph{Proceedings of the IEEE/CVF Conference on Computer Vision and Pattern Recognition}, pages 11909--11919, 2023.

\bibitem[Sun et~al.(2024)Sun, Yang, Zhang, Cheng, and Hou]{sun2023corrmatch}
Boyuan Sun, Yuqi Yang, Le Zhang, Ming-Ming Cheng, and Qibin Hou.
\newblock Corrmatch: Label propagation via correlation matching for semi-supervised semantic segmentation.
\newblock \emph{IEEE Computer Vision and Pattern Recognition (CVPR)}, 2024.

\bibitem[Sun et~al.(2019)Sun, Myers, Vondrick, Murphy, and Schmid]{sun2019videobert}
Chen Sun, Austin Myers, Carl Vondrick, Kevin Murphy, and Cordelia Schmid.
\newblock Videobert: A joint model for video and language representation learning.
\newblock In \emph{Proceedings of the IEEE/CVF international conference on computer vision}, pages 7464--7473, 2019.

\bibitem[Sun et~al.(2023{\natexlab{a}})Sun, Zhou, Ye, and Zhan]{sun2023pilot}
Hai-Long Sun, Da-Wei Zhou, Han-Jia Ye, and De-Chuan Zhan.
\newblock Pilot: A pre-trained model-based continual learning toolbox.
\newblock \emph{arXiv preprint arXiv:2309.07117}, 2023{\natexlab{a}}.

\bibitem[Sun et~al.(2023{\natexlab{b}})Sun, Fang, Wu, Wang, and Cao]{sun2023eva}
Quan Sun, Yuxin Fang, Ledell Wu, Xinlong Wang, and Yue Cao.
\newblock Eva-clip: Improved training techniques for clip at scale.
\newblock \emph{arXiv preprint arXiv:2303.15389}, 2023{\natexlab{b}}.

\bibitem[Tan and Bansal(2019)]{tan2019lxmert}
Hao Tan and Mohit Bansal.
\newblock Lxmert: Learning cross-modality encoder representations from transformers.
\newblock \emph{arXiv preprint arXiv:1908.07490}, 2019.

\bibitem[Tao et~al.(2020{\natexlab{a}})Tao, Chang, Hong, Wei, and Gong]{tao2020topology}
Xiaoyu Tao, Xinyuan Chang, Xiaopeng Hong, Xing Wei, and Yihong Gong.
\newblock Topology-preserving class-incremental learning.
\newblock In \emph{ECCV}, 2020{\natexlab{a}}.

\bibitem[Tao et~al.(2020{\natexlab{b}})Tao, Hong, Chang, Dong, Wei, and Gong]{tao2020few}
Xiaoyu Tao, Xiaopeng Hong, Xinyuan Chang, Songlin Dong, Xing Wei, and Yihong Gong.
\newblock Few-shot class-incremental learning.
\newblock In \emph{Proceedings of the IEEE/CVF Conference on Computer Vision and Pattern Recognition}, pages 12183--12192, 2020{\natexlab{b}}.

\bibitem[Thengane et~al.(2022)Thengane, Khan, Hayat, and Khan]{continual-clip}
Vishal Thengane, Salman Khan, Munawar Hayat, and Fahad Khan.
\newblock Clip model is an efficient continual learner.
\newblock \emph{arXiv preprint arXiv:2210.03114}, 2022.

\bibitem[Touvron et~al.(2023)Touvron, Lavril, Izacard, Martinet, Lachaux, Lacroix, Rozi{\`e}re, Goyal, Hambro, Azhar, et~al.]{touvron2023llama}
Hugo Touvron, Thibaut Lavril, Gautier Izacard, Xavier Martinet, Marie-Anne Lachaux, Timoth{\'e}e Lacroix, Baptiste Rozi{\`e}re, Naman Goyal, Eric Hambro, Faisal Azhar, et~al.
\newblock Llama: Open and efficient foundation language models.
\newblock \emph{arXiv preprint arXiv:2302.13971}, 2023.

\bibitem[Tsimpoukelli et~al.(2021)Tsimpoukelli, Menick, Cabi, Eslami, Vinyals, and Hill]{tsimpoukelli2021multimodal}
Maria Tsimpoukelli, Jacob~L Menick, Serkan Cabi, SM Eslami, Oriol Vinyals, and Felix Hill.
\newblock Multimodal few-shot learning with frozen language models.
\newblock \emph{Advances in Neural Information Processing Systems}, 34:\penalty0 200--212, 2021.

\bibitem[Van~de Ven and Tolias(2019)]{van2019three}
Gido~M Van~de Ven and Andreas~S Tolias.
\newblock Three scenarios for continual learning.
\newblock \emph{NIPS Workshops}, 2019.

\bibitem[Wang et~al.(2023{\natexlab{a}})Wang, Zhang, Su, and Zhu]{wang2023comprehensive}
Liyuan Wang, Xingxing Zhang, Hang Su, and Jun Zhu.
\newblock A comprehensive survey of continual learning: Theory, method and application.
\newblock \emph{arXiv preprint arXiv:2302.00487}, 2023{\natexlab{a}}.

\bibitem[Wang et~al.(2024)Wang, Xie, Zhang, Huang, Su, and Zhu]{wang2024hierarchical}
Liyuan Wang, Jingyi Xie, Xingxing Zhang, Mingyi Huang, Hang Su, and Jun Zhu.
\newblock Hierarchical decomposition of prompt-based continual learning: Rethinking obscured sub-optimality.
\newblock \emph{Advances in Neural Information Processing Systems}, 36, 2024.

\bibitem[Wang et~al.(2023{\natexlab{b}})Wang, Chen, Chen, Wu, Zhu, Zeng, Luo, Lu, Zhou, Qiao, et~al.]{wang2023visionllm}
Wenhai Wang, Zhe Chen, Xiaokang Chen, Jiannan Wu, Xizhou Zhu, Gang Zeng, Ping Luo, Tong Lu, Jie Zhou, Yu Qiao, et~al.
\newblock Visionllm: Large language model is also an open-ended decoder for vision-centric tasks.
\newblock \emph{arXiv preprint arXiv:2305.11175}, 2023{\natexlab{b}}.

\bibitem[Wang et~al.(2023{\natexlab{c}})Wang, Chen, Qian, Gao, Wei, Wang, Tian, and Gao]{wang2023large}
Xiao Wang, Guangyao Chen, Guangwu Qian, Pengcheng Gao, Xiao-Yong Wei, Yaowei Wang, Yonghong Tian, and Wen Gao.
\newblock Large-scale multi-modal pre-trained models: A comprehensive survey.
\newblock \emph{Machine Intelligence Research}, 20\penalty0 (4):\penalty0 447--482, 2023{\natexlab{c}}.

\bibitem[Wang et~al.(2022{\natexlab{a}})Wang, Zhang, Ebrahimi, Sun, Zhang, Lee, Ren, Su, Perot, Dy, et~al.]{wang2022dualprompt}
Zifeng Wang, Zizhao Zhang, Sayna Ebrahimi, Ruoxi Sun, Han Zhang, Chen-Yu Lee, Xiaoqi Ren, Guolong Su, Vincent Perot, Jennifer Dy, et~al.
\newblock Dualprompt: Complementary prompting for rehearsal-free continual learning.
\newblock In \emph{ECCV}, 2022{\natexlab{a}}.

\bibitem[Wang et~al.(2022{\natexlab{b}})Wang, Zhang, Lee, Zhang, Sun, Ren, Su, Perot, Dy, and Pfister]{l2p}
Zifeng Wang, Zizhao Zhang, Chen{-}Yu Lee, Han Zhang, Ruoxi Sun, Xiaoqi Ren, Guolong Su, Vincent Perot, Jennifer~G. Dy, and Tomas Pfister.
\newblock Learning to prompt for continual learning.
\newblock \emph{CVPR}, 2022{\natexlab{b}}.

\bibitem[Wu et~al.(2022)Wu, Swaminathan, Li, Ravichandran, Vasconcelos, Bhotika, and Soatto]{wu2022class}
Tz-Ying Wu, Gurumurthy Swaminathan, Zhizhong Li, Avinash Ravichandran, Nuno Vasconcelos, Rahul Bhotika, and Stefano Soatto.
\newblock Class-incremental learning with strong pre-trained models.
\newblock In \emph{Proceedings of the IEEE/CVF Conference on Computer Vision and Pattern Recognition}, pages 9601--9610, 2022.

\bibitem[Wu et~al.(2019{\natexlab{a}})Wu, Chen, Wang, Ye, Liu, Guo, and Fu]{BiC}
Yue Wu, Yinpeng Chen, Lijuan Wang, Yuancheng Ye, Zicheng Liu, Yandong Guo, and Yun Fu.
\newblock Large scale incremental learning.
\newblock In \emph{CVPR}, 2019{\natexlab{a}}.

\bibitem[Wu et~al.(2019{\natexlab{b}})Wu, Chen, Wang, Ye, Liu, Guo, and Fu]{wu2019large}
Yue Wu, Yinpeng Chen, Lijuan Wang, Yuancheng Ye, Zicheng Liu, Yandong Guo, and Yun Fu.
\newblock Large scale incremental learning.
\newblock In \emph{ICCV}, 2019{\natexlab{b}}.

\bibitem[Yan et~al.(2021)Yan, Xie, and He]{yan2021dynamically}
Shipeng Yan, Jiangwei Xie, and Xuming He.
\newblock Der: Dynamically expandable representation for class incremental learning.
\newblock In \emph{CVPR}, 2021.

\bibitem[Yang et~al.(2023)Yang, Cui, Xu, Zhong, Zheng, and Wang]{yang2023continual}
Yang Yang, Zhiying Cui, Junjie Xu, Changhong Zhong, Wei-Shi Zheng, and Ruixuan Wang.
\newblock Continual learning with bayesian model based on a fixed pre-trained feature extractor.
\newblock \emph{Visual Intelligence}, 1\penalty0 (1):\penalty0 5, 2023.

\bibitem[Yu et~al.(2020)Yu, Twardowski, Liu, Herranz, Wang, Cheng, Jui, and Weijer]{yu2020semantic}
Lu Yu, Bartlomiej Twardowski, Xialei Liu, Luis Herranz, Kai Wang, Yongmei Cheng, Shangling Jui, and Joost van~de Weijer.
\newblock Semantic drift compensation for class-incremental learning.
\newblock In \emph{Proceedings of the IEEE/CVF conference on computer vision and pattern recognition}, pages 6982--6991, 2020.

\bibitem[Zhang et~al.(2021)Zhang, Song, Lin, Zheng, Pan, and Xu]{zhang2021few}
Chi Zhang, Nan Song, Guosheng Lin, Yun Zheng, Pan Pan, and Yinghui Xu.
\newblock Few-shot incremental learning with continually evolved classifiers.
\newblock In \emph{Proceedings of the IEEE/CVF conference on computer vision and pattern recognition}, pages 12455--12464, 2021.

\bibitem[Zhang et~al.(2022)Zhang, Xiao, Liu, Chen, and Cheng]{zhang2022representation}
Chang-Bin Zhang, Jia-Wen Xiao, Xialei Liu, Ying-Cong Chen, and Ming-Ming Cheng.
\newblock Representation compensation networks for continual semantic segmentation.
\newblock In \emph{Proceedings of the IEEE/CVF Conference on Computer Vision and Pattern Recognition}, pages 7053--7064, 2022.

\bibitem[Zhang et~al.(2023)Zhang, Wang, Kang, Chen, and Wei]{zhang2023slca}
Gengwei Zhang, Liyuan Wang, Guoliang Kang, Ling Chen, and Yunchao Wei.
\newblock Slca: Slow learner with classifier alignment for continual learning on a pre-trained model.
\newblock \emph{arXiv preprint arXiv:2303.05118}, 2023.

\bibitem[Zhao et~al.(2020)Zhao, Xiao, Gan, Zhang, and Xia]{zhao2020maintaining}
Bowen Zhao, Xi Xiao, Guojun Gan, Bin Zhang, and Shu-Tao Xia.
\newblock Maintaining discrimination and fairness in class incremental learning.
\newblock In \emph{CVPR}, 2020.

\bibitem[Zhao et~al.(2021)Zhao, Fu, Kang, Tian, Wu, and Li]{zhao2021mgsvf}
Hanbin Zhao, Yongjian Fu, Mintong Kang, Qi Tian, Fei Wu, and Xi Li.
\newblock Mgsvf: Multi-grained slow vs. fast framework for few-shot class-incremental learning.
\newblock \emph{IEEE Transactions on Pattern Analysis and Machine Intelligence}, 2021.

\bibitem[Zhou et~al.(2023{\natexlab{a}})Zhou, Wang, Qi, Ye, Zhan, and Liu]{zhou2023deep}
Da-Wei Zhou, Qi-Wei Wang, Zhi-Hong Qi, Han-Jia Ye, De-Chuan Zhan, and Ziwei Liu.
\newblock Deep class-incremental learning: A survey.
\newblock \emph{arXiv preprint arXiv:2302.03648}, 2023{\natexlab{a}}.

\bibitem[Zhou et~al.(2023{\natexlab{b}})Zhou, Ye, Zhan, and Liu]{zhou2023revisiting}
Da-Wei Zhou, Han-Jia Ye, De-Chuan Zhan, and Ziwei Liu.
\newblock Revisiting class-incremental learning with pre-trained models: Generalizability and adaptivity are all you need, 2023{\natexlab{b}}.

\bibitem[Zhou et~al.(2023{\natexlab{c}})Zhou, Zhang, Ning, Ye, Zhan, and Liu]{zhou2023learning}
Da-Wei Zhou, Yuanhan Zhang, Jingyi Ning, Han-Jia Ye, De-Chuan Zhan, and Ziwei Liu.
\newblock Learning without forgetting for vision-language models.
\newblock \emph{arXiv preprint arXiv:2305.19270}, 2023{\natexlab{c}}.

\bibitem[Zhou et~al.(2023{\natexlab{d}})Zhou, Yu, Zhang, Wu, Wang, and Wang]{zhou2023regionblip}
Qiang Zhou, Chaohui Yu, Shaofeng Zhang, Sitong Wu, Zhibing Wang, and Fan Wang.
\newblock Regionblip: A unified multi-modal pre-training framework for holistic and regional comprehension.
\newblock \emph{arXiv preprint arXiv:2308.02299}, 2023{\natexlab{d}}.

\bibitem[Zhu et~al.(2023)Zhu, Chen, Shen, Li, and Elhoseiny]{zhu2023minigpt}
Deyao Zhu, Jun Chen, Xiaoqian Shen, Xiang Li, and Mohamed Elhoseiny.
\newblock Minigpt-4: Enhancing vision-language understanding with advanced large language models.
\newblock \emph{arXiv preprint arXiv:2304.10592}, 2023.

\bibitem[Zhu et~al.(2021{\natexlab{a}})Zhu, Zhang, Wang, Yin, and Liu]{zhu2021prototype}
Fei Zhu, Xu-Yao Zhang, Chuang Wang, Fei Yin, and Cheng-Lin Liu.
\newblock Prototype augmentation and self-supervision for incremental learning.
\newblock In \emph{Proceedings of the IEEE/CVF Conference on Computer Vision and Pattern Recognition}, pages 5871--5880, 2021{\natexlab{a}}.

\bibitem[Zhu et~al.(2021{\natexlab{b}})Zhu, Cao, Zhai, Cheng, and Zha]{zhu2021self}
Kai Zhu, Yang Cao, Wei Zhai, Jie Cheng, and Zheng-Jun Zha.
\newblock Self-promoted prototype refinement for few-shot class-incremental learning.
\newblock In \emph{Proceedings of the IEEE/CVF conference on computer vision and pattern recognition}, pages 6801--6810, 2021{\natexlab{b}}.

\end{thebibliography}


\begin{thebibliography}{63}
\providecommand{\natexlab}[1]{#1}
\providecommand{\url}[1]{\texttt{#1}}
\expandafter\ifx\csname urlstyle\endcsname\relax
  \providecommand{\doi}[1]{doi: #1}\else
  \providecommand{\doi}{doi: \begingroup \urlstyle{rm}\Url}\fi

\bibitem[Ahn et~al.(2021)Ahn, Kwak, Lim, Bang, Kim, and Moon]{ahn2021ss}
Hongjoon Ahn, Jihwan Kwak, Subin Lim, Hyeonsu Bang, Hyojun Kim, and Taesup Moon.
\newblock Ss-il: Separated softmax for incremental learning.
\newblock In \emph{ICCV}, 2021.

\bibitem[Aljundi et~al.(2017)Aljundi, Chakravarty, and Tuytelaars]{aljundi2017expert}
Rahaf Aljundi, Punarjay Chakravarty, and Tinne Tuytelaars.
\newblock Expert gate: Lifelong learning with a network of experts.
\newblock In \emph{Proceedings of the IEEE conference on computer vision and pattern recognition}, pages 3366--3375, 2017.

\bibitem[Ben-Younes et~al.(2017)Ben-Younes, Cadene, Cord, and Thome]{ben2017mutan}
Hedi Ben-Younes, R{\'e}mi Cadene, Matthieu Cord, and Nicolas Thome.
\newblock Mutan: Multimodal tucker fusion for visual question answering.
\newblock In \emph{Proceedings of the IEEE international conference on computer vision}, pages 2612--2620, 2017.

\bibitem[Caron et~al.(2021)Caron, Touvron, Misra, J{\'e}gou, Mairal, Bojanowski, and Joulin]{caron2021emerging}
Mathilde Caron, Hugo Touvron, Ishan Misra, Herv{\'e} J{\'e}gou, Julien Mairal, Piotr Bojanowski, and Armand Joulin.
\newblock Emerging properties in self-supervised vision transformers.
\newblock In \emph{Proceedings of the IEEE/CVF international conference on computer vision}, pages 9650--9660, 2021.

\bibitem[Castro et~al.(2018)Castro, Mar{\'\i}n-Jim{\'e}nez, Guil, Schmid, and Alahari]{castro2018end}
Francisco~M Castro, Manuel~J Mar{\'\i}n-Jim{\'e}nez, Nicol{\'a}s Guil, Cordelia Schmid, and Karteek Alahari.
\newblock End-to-end incremental learning.
\newblock In \emph{Proceedings of the European conference on computer vision (ECCV)}, pages 233--248, 2018.

\bibitem[Chen et~al.(2020)Chen, Fan, Girshick, and He]{chen2020improved}
Xinlei Chen, Haoqi Fan, Ross Girshick, and Kaiming He.
\newblock Improved baselines with momentum contrastive learning.
\newblock \emph{arXiv preprint arXiv:2003.04297}, 2020.

\bibitem[Chen* et~al.(2021)Chen*, Xie*, and He]{chen2021mocov3}
Xinlei Chen*, Saining Xie*, and Kaiming He.
\newblock An empirical study of training self-supervised vision transformers.
\newblock \emph{arXiv preprint arXiv:2104.02057}, 2021.

\bibitem[Chi et~al.(2022)Chi, Gu, Liu, Wang, Yu, and Tang]{chi2022metafscil}
Zhixiang Chi, Li Gu, Huan Liu, Yang Wang, Yuanhao Yu, and Jin Tang.
\newblock Metafscil: A meta-learning approach for few-shot class incremental learning.
\newblock In \emph{Proceedings of the IEEE/CVF conference on computer vision and pattern recognition}, pages 14166--14175, 2022.

\bibitem[Chiang et~al.(2023)Chiang, Li, Lin, Sheng, Wu, Zhang, Zheng, Zhuang, Zhuang, Gonzalez, et~al.]{chiang2023vicuna}
Wei-Lin Chiang, Zhuohan Li, Zi Lin, Ying Sheng, Zhanghao Wu, Hao Zhang, Lianmin Zheng, Siyuan Zhuang, Yonghao Zhuang, Joseph~E Gonzalez, et~al.
\newblock Vicuna: An open-source chatbot impressing gpt-4 with 90\%* chatgpt quality.
\newblock \emph{See https://vicuna. lmsys. org (accessed 14 April 2023)}, 2023.

\bibitem[Delange et~al.(2021)Delange, Aljundi, Masana, Parisot, Jia, Leonardis, Slabaugh, and Tuytelaars]{delange2021continual}
Matthias Delange, Rahaf Aljundi, Marc Masana, Sarah Parisot, Xu Jia, Ales Leonardis, Greg Slabaugh, and Tinne Tuytelaars.
\newblock A continual learning survey: Defying forgetting in classification tasks.
\newblock \emph{TPAMI}, 2021.

\bibitem[Douillard et~al.(2020)Douillard, Cord, Ollion, Robert, and Valle]{douillard2020podnet}
Arthur Douillard, Matthieu Cord, Charles Ollion, Thomas Robert, and Eduardo Valle.
\newblock Podnet: Pooled outputs distillation for small-tasks incremental learning.
\newblock In \emph{ECCV}, 2020.

\bibitem[Douillard et~al.(2022)Douillard, Ram{\'e}, Couairon, and Cord]{douillard2022dytox}
Arthur Douillard, Alexandre Ram{\'e}, Guillaume Couairon, and Matthieu Cord.
\newblock Dytox: Transformers for continual learning with dynamic token expansion.
\newblock In \emph{CVPR}, 2022.

\bibitem[Fernando et~al.(2017)Fernando, Banarse, Blundell, Zwols, Ha, Rusu, Pritzel, and Wierstra]{fernando2017pathnet}
Chrisantha Fernando, Dylan Banarse, Charles Blundell, Yori Zwols, David Ha, Andrei~A Rusu, Alexander Pritzel, and Daan Wierstra.
\newblock Pathnet: Evolution channels gradient descent in super neural networks.
\newblock \emph{arXiv preprint arXiv:1701.08734}, 2017.

\bibitem[Gao and Liu(2023)]{gao2023ddgr}
Rui Gao and Weiwei Liu.
\newblock Ddgr: Continual learning with deep diffusion-based generative replay.
\newblock In \emph{ICML}, 2023.

\bibitem[Hayes et~al.(2020)Hayes, Kafle, Shrestha, Acharya, and Kanan]{hayes2020remind}
Tyler~L Hayes, Kushal Kafle, Robik Shrestha, Manoj Acharya, and Christopher Kanan.
\newblock Remind your neural network to prevent catastrophic forgetting.
\newblock In \emph{ECCV}, 2020.

\bibitem[He et~al.(2020)He, Fan, Wu, Xie, and Girshick]{he2020momentum}
Kaiming He, Haoqi Fan, Yuxin Wu, Saining Xie, and Ross Girshick.
\newblock Momentum contrast for unsupervised visual representation learning.
\newblock In \emph{Proceedings of the IEEE/CVF conference on computer vision and pattern recognition}, pages 9729--9738, 2020.

\bibitem[Hendrycks et~al.(2021)Hendrycks, Basart, Mu, Kadavath, Wang, Dorundo, Desai, Zhu, Parajuli, Guo, Song, Steinhardt, and Gilmer]{hendrycks2021many}
Dan Hendrycks, Steven Basart, Norman Mu, Saurav Kadavath, Frank Wang, Evan Dorundo, Rahul Desai, Tyler Zhu, Samyak Parajuli, Mike Guo, Dawn Song, Jacob Steinhardt, and Justin Gilmer.
\newblock The many faces of robustness: A critical analysis of out-of-distribution generalization.
\newblock \emph{ICCV}, 2021.

\bibitem[Hou et~al.(2019)Hou, Pan, Loy, Wang, and Lin]{UCIR_2019_CVPR}
Saihui Hou, Xinyu Pan, Chen~Change Loy, Zilei Wang, and Dahua Lin.
\newblock Learning a unified classifier incrementally via rebalancing.
\newblock In \emph{CVPR}, 2019.

\bibitem[Hu et~al.(2021)Hu, Tang, Miao, Hua, and Zhang]{hu2021distilling}
Xinting Hu, Kaihua Tang, Chunyan Miao, Xian-Sheng Hua, and Hanwang Zhang.
\newblock Distilling causal effect of data in class-incremental learning.
\newblock In \emph{CVPR}, 2021.

\bibitem[Kirkpatrick et~al.(2017)Kirkpatrick, Pascanu, Rabinowitz, Veness, Desjardins, Rusu, Milan, Quan, Ramalho, Grabska-Barwinska, et~al.]{kirkpatrick2017overcoming}
James Kirkpatrick, Razvan Pascanu, Neil Rabinowitz, Joel Veness, Guillaume Desjardins, Andrei~A Rusu, Kieran Milan, John Quan, Tiago Ramalho, Agnieszka Grabska-Barwinska, et~al.
\newblock Overcoming catastrophic forgetting in neural networks.
\newblock \emph{Proceedings of the national academy of sciences}, 114\penalty0 (13):\penalty0 3521--3526, 2017.

\bibitem[Li et~al.(2022)Li, Li, Xiong, and Hoi]{li2022blip}
Junnan Li, Dongxu Li, Caiming Xiong, and Steven Hoi.
\newblock Blip: Bootstrapping language-image pre-training for unified vision-language understanding and generation.
\newblock In \emph{International Conference on Machine Learning}, pages 12888--12900. PMLR, 2022.

\bibitem[Li et~al.(2023)Li, Li, Savarese, and Hoi]{li2023blip}
Junnan Li, Dongxu Li, Silvio Savarese, and Steven Hoi.
\newblock Blip-2: Bootstrapping language-image pre-training with frozen image encoders and large language models.
\newblock \emph{arXiv preprint arXiv:2301.12597}, 2023.

\bibitem[Li et~al.(2019)Li, Zhang, Zhang, Huang, He, Lyu, and Gao]{li2019object}
Wenbo Li, Pengchuan Zhang, Lei Zhang, Qiuyuan Huang, Xiaodong He, Siwei Lyu, and Jianfeng Gao.
\newblock Object-driven text-to-image synthesis via adversarial training.
\newblock In \emph{Proceedings of the IEEE/CVF Conference on Computer Vision and Pattern Recognition}, pages 12174--12182, 2019.

\bibitem[Li and Hoiem(2017)]{li2017learning}
Zhizhong Li and Derek Hoiem.
\newblock Learning without forgetting.
\newblock \emph{IEEE transactions on pattern analysis and machine intelligence}, 40\penalty0 (12):\penalty0 2935--2947, 2017.

\bibitem[Li and Hoiem(2018)]{li2018learning}
Zhizhong Li and Derek Hoiem.
\newblock Learning without forgetting.
\newblock \emph{TPAMI}, 2018.

\bibitem[Liu et~al.(2022)Liu, Gu, Chi, Wang, Yu, Chen, and Tang]{liu2022few}
Huan Liu, Li Gu, Zhixiang Chi, Yang Wang, Yuanhao Yu, Jun Chen, and Jin Tang.
\newblock Few-shot class-incremental learning via entropy-regularized data-free replay.
\newblock In \emph{European Conference on Computer Vision}, pages 146--162. Springer, 2022.

\bibitem[Liu et~al.(2023)Liu, Li, Wu, and Lee]{liu2023visual}
Haotian Liu, Chunyuan Li, Qingyang Wu, and Yong~Jae Lee.
\newblock Visual instruction tuning.
\newblock \emph{arXiv preprint arXiv:2304.08485}, 2023.

\bibitem[Liu et~al.(2021{\natexlab{a}})Liu, Chen, Guo, Zhu, and Liu]{liu2021cptr}
Wei Liu, Sihan Chen, Longteng Guo, Xinxin Zhu, and Jing Liu.
\newblock Cptr: Full transformer network for image captioning.
\newblock \emph{arXiv preprint arXiv:2101.10804}, 2021{\natexlab{a}}.

\bibitem[Liu et~al.(2018)Liu, Masana, Herranz, Van~de Weijer, Lopez, and Bagdanov]{liu2018rotate}
Xialei Liu, Marc Masana, Luis Herranz, Joost Van~de Weijer, Antonio~M Lopez, and Andrew~D Bagdanov.
\newblock Rotate your networks: Better weight consolidation and less catastrophic forgetting.
\newblock In \emph{ICPR}, 2018.

\bibitem[Liu et~al.(2021{\natexlab{b}})Liu, Schiele, and Sun]{liu2021adaptive}
Yaoyao Liu, Bernt Schiele, and Qianru Sun.
\newblock Adaptive aggregation networks for class-incremental learning.
\newblock In \emph{CVPR}, 2021{\natexlab{b}}.

\bibitem[Lu et~al.(2019)Lu, Batra, Parikh, and Lee]{lu2019vilbert}
Jiasen Lu, Dhruv Batra, Devi Parikh, and Stefan Lee.
\newblock Vilbert: Pretraining task-agnostic visiolinguistic representations for vision-and-language tasks.
\newblock \emph{Advances in neural information processing systems}, 32, 2019.

\bibitem[Mallya et~al.(2018)Mallya, Davis, and Lazebnik]{mallya2018piggyback}
Arun Mallya, Dillon Davis, and Svetlana Lazebnik.
\newblock Piggyback: Adapting a single network to multiple tasks by learning to mask weights.
\newblock In \emph{ECCV}, 2018.

\bibitem[McCloskey and Cohen(1989)]{mccloskey1989catastrophic}
Michael McCloskey and Neal~J Cohen.
\newblock Catastrophic interference in connectionist networks: The sequential learning problem.
\newblock In \emph{Psychology of learning and motivation}, pages 109--165. Elsevier, 1989.

\bibitem[OpenAI(2023)]{openai2023gpt}
R OpenAI.
\newblock Gpt-4 technical report.
\newblock \emph{arXiv}, pages 2303--08774, 2023.

\bibitem[Radford et~al.(2021)Radford, Kim, Hallacy, Ramesh, Goh, Agarwal, Sastry, Askell, Mishkin, Clark, et~al.]{clip}
Alec Radford, Jong~Wook Kim, Chris Hallacy, Aditya Ramesh, Gabriel Goh, Sandhini Agarwal, Girish Sastry, Amanda Askell, Pamela Mishkin, Jack Clark, et~al.
\newblock Learning transferable visual models from natural language supervision.
\newblock In \emph{ICML}, 2021.

\bibitem[Rajasegaran et~al.(2019)Rajasegaran, Hayat, Khan, Khan, and Shao]{rajasegaran2019random}
Jathushan Rajasegaran, Munawar Hayat, Salman Khan, Fahad~Shahbaz Khan, and Ling Shao.
\newblock Random path selection for incremental learning.
\newblock In \emph{NIPS}, 2019.

\bibitem[Rebuffi et~al.(2017)Rebuffi, Kolesnikov, Sperl, and Lampert]{rebuffi2017icarl}
Sylvestre-Alvise Rebuffi, Alexander Kolesnikov, Georg Sperl, and Christoph~H Lampert.
\newblock icarl: Incremental classifier and representation learning.
\newblock In \emph{CVPR}, 2017.

\bibitem[Russakovsky et~al.(2015)Russakovsky, Deng, Su, Krause, Satheesh, Ma, Huang, Karpathy, Khosla, Bernstein, et~al.]{russakovsky2015imagenet}
Olga Russakovsky, Jia Deng, Hao Su, Jonathan Krause, Sanjeev Satheesh, Sean Ma, Zhiheng Huang, Andrej Karpathy, Aditya Khosla, Michael Bernstein, et~al.
\newblock Imagenet large scale visual recognition challenge.
\newblock \emph{International journal of computer vision}, 115:\penalty0 211--252, 2015.

\bibitem[Schwarz et~al.(2018)Schwarz, Czarnecki, Luketina, Grabska-Barwinska, Whye~Teh, Pascanu, and Hadsell]{schwarz2018progress}
Jonathan Schwarz, Wojciech Czarnecki, Jelena Luketina, Agnieszka Grabska-Barwinska, Yee Whye~Teh, Razvan Pascanu, and Raia Hadsell.
\newblock Progress \& compress: A scalable framework for continual learning.
\newblock In \emph{ICML}, 2018.

\bibitem[Serra et~al.(2018)Serra, Suris, Miron, and Karatzoglou]{serra2018overcoming}
Joan Serra, Didac Suris, Marius Miron, and Alexandros Karatzoglou.
\newblock Overcoming catastrophic forgetting with hard attention to the task.
\newblock In \emph{ICML}, 2018.

\bibitem[Shi et~al.(2021)Shi, Chen, Zhang, Zhan, and Wu]{shi2021overcoming}
Guangyuan Shi, Jiaxin Chen, Wenlong Zhang, Li-Ming Zhan, and Xiao-Ming Wu.
\newblock Overcoming catastrophic forgetting in incremental few-shot learning by finding flat minima.
\newblock \emph{Advances in neural information processing systems}, 34:\penalty0 6747--6761, 2021.

\bibitem[Shin et~al.(2017)Shin, Lee, Kim, and Kim]{shin2017continual}
Hanul Shin, Jung~Kwon Lee, Jaehong Kim, and Jiwon Kim.
\newblock Continual learning with deep generative replay.
\newblock \emph{Advances in neural information processing systems}, 30, 2017.

\bibitem[Smith et~al.(2023)Smith, Karlinsky, Gutta, Cascante-Bonilla, Kim, Arbelle, Panda, Feris, and Kira]{smith2023coda}
James~Seale Smith, Leonid Karlinsky, Vyshnavi Gutta, Paola Cascante-Bonilla, Donghyun Kim, Assaf Arbelle, Rameswar Panda, Rogerio Feris, and Zsolt Kira.
\newblock Coda-prompt: Continual decomposed attention-based prompting for rehearsal-free continual learning.
\newblock In \emph{Proceedings of the IEEE/CVF Conference on Computer Vision and Pattern Recognition}, pages 11909--11919, 2023.

\bibitem[Sun et~al.(2019)Sun, Myers, Vondrick, Murphy, and Schmid]{sun2019videobert}
Chen Sun, Austin Myers, Carl Vondrick, Kevin Murphy, and Cordelia Schmid.
\newblock Videobert: A joint model for video and language representation learning.
\newblock In \emph{Proceedings of the IEEE/CVF international conference on computer vision}, pages 7464--7473, 2019.

\bibitem[Tan and Bansal(2019)]{tan2019lxmert}
Hao Tan and Mohit Bansal.
\newblock Lxmert: Learning cross-modality encoder representations from transformers.
\newblock \emph{arXiv preprint arXiv:1908.07490}, 2019.

\bibitem[Tao et~al.(2020{\natexlab{a}})Tao, Chang, Hong, Wei, and Gong]{tao2020topology}
Xiaoyu Tao, Xinyuan Chang, Xiaopeng Hong, Xing Wei, and Yihong Gong.
\newblock Topology-preserving class-incremental learning.
\newblock In \emph{ECCV}, 2020{\natexlab{a}}.

\bibitem[Tao et~al.(2020{\natexlab{b}})Tao, Hong, Chang, Dong, Wei, and Gong]{tao2020few}
Xiaoyu Tao, Xiaopeng Hong, Xinyuan Chang, Songlin Dong, Xing Wei, and Yihong Gong.
\newblock Few-shot class-incremental learning.
\newblock In \emph{Proceedings of the IEEE/CVF Conference on Computer Vision and Pattern Recognition}, pages 12183--12192, 2020{\natexlab{b}}.

\bibitem[Thengane et~al.(2022)Thengane, Khan, Hayat, and Khan]{continual-clip}
Vishal Thengane, Salman Khan, Munawar Hayat, and Fahad Khan.
\newblock Clip model is an efficient continual learner.
\newblock \emph{arXiv preprint arXiv:2210.03114}, 2022.

\bibitem[Touvron et~al.(2023)Touvron, Lavril, Izacard, Martinet, Lachaux, Lacroix, Rozi{\`e}re, Goyal, Hambro, Azhar, et~al.]{touvron2023llama}
Hugo Touvron, Thibaut Lavril, Gautier Izacard, Xavier Martinet, Marie-Anne Lachaux, Timoth{\'e}e Lacroix, Baptiste Rozi{\`e}re, Naman Goyal, Eric Hambro, Faisal Azhar, et~al.
\newblock Llama: Open and efficient foundation language models.
\newblock \emph{arXiv preprint arXiv:2302.13971}, 2023.

\bibitem[Van~de Ven and Tolias(2019)]{van2019three}
Gido~M Van~de Ven and Andreas~S Tolias.
\newblock Three scenarios for continual learning.
\newblock \emph{NIPS Workshops}, 2019.

\bibitem[Wang et~al.(2022{\natexlab{a}})Wang, Zhang, Ebrahimi, Sun, Zhang, Lee, Ren, Su, Perot, Dy, et~al.]{wang2022dualprompt}
Zifeng Wang, Zizhao Zhang, Sayna Ebrahimi, Ruoxi Sun, Han Zhang, Chen-Yu Lee, Xiaoqi Ren, Guolong Su, Vincent Perot, Jennifer Dy, et~al.
\newblock Dualprompt: Complementary prompting for rehearsal-free continual learning.
\newblock In \emph{ECCV}, 2022{\natexlab{a}}.

\bibitem[Wang et~al.(2022{\natexlab{b}})Wang, Zhang, Lee, Zhang, Sun, Ren, Su, Perot, Dy, and Pfister]{l2p}
Zifeng Wang, Zizhao Zhang, Chen{-}Yu Lee, Han Zhang, Ruoxi Sun, Xiaoqi Ren, Guolong Su, Vincent Perot, Jennifer~G. Dy, and Tomas Pfister.
\newblock Learning to prompt for continual learning.
\newblock \emph{CVPR}, 2022{\natexlab{b}}.

\bibitem[Wu et~al.(2022)Wu, Swaminathan, Li, Ravichandran, Vasconcelos, Bhotika, and Soatto]{wu2022class}
Tz-Ying Wu, Gurumurthy Swaminathan, Zhizhong Li, Avinash Ravichandran, Nuno Vasconcelos, Rahul Bhotika, and Stefano Soatto.
\newblock Class-incremental learning with strong pre-trained models.
\newblock In \emph{Proceedings of the IEEE/CVF Conference on Computer Vision and Pattern Recognition}, pages 9601--9610, 2022.

\bibitem[Wu et~al.(2019{\natexlab{a}})Wu, Chen, Wang, Ye, Liu, Guo, and Fu]{BiC}
Yue Wu, Yinpeng Chen, Lijuan Wang, Yuancheng Ye, Zicheng Liu, Yandong Guo, and Yun Fu.
\newblock Large scale incremental learning.
\newblock In \emph{CVPR}, 2019{\natexlab{a}}.

\bibitem[Wu et~al.(2019{\natexlab{b}})Wu, Chen, Wang, Ye, Liu, Guo, and Fu]{wu2019large}
Yue Wu, Yinpeng Chen, Lijuan Wang, Yuancheng Ye, Zicheng Liu, Yandong Guo, and Yun Fu.
\newblock Large scale incremental learning.
\newblock In \emph{ICCV}, 2019{\natexlab{b}}.

\bibitem[Yan et~al.(2021)Yan, Xie, and He]{yan2021dynamically}
Shipeng Yan, Jiangwei Xie, and Xuming He.
\newblock Der: Dynamically expandable representation for class incremental learning.
\newblock In \emph{CVPR}, 2021.

\bibitem[Zhang et~al.(2021)Zhang, Song, Lin, Zheng, Pan, and Xu]{zhang2021few}
Chi Zhang, Nan Song, Guosheng Lin, Yun Zheng, Pan Pan, and Yinghui Xu.
\newblock Few-shot incremental learning with continually evolved classifiers.
\newblock In \emph{Proceedings of the IEEE/CVF conference on computer vision and pattern recognition}, pages 12455--12464, 2021.

\bibitem[Zhang et~al.(2023)Zhang, Wang, Kang, Chen, and Wei]{zhang2023slca}
Gengwei Zhang, Liyuan Wang, Guoliang Kang, Ling Chen, and Yunchao Wei.
\newblock Slca: Slow learner with classifier alignment for continual learning on a pre-trained model.
\newblock \emph{arXiv preprint arXiv:2303.05118}, 2023.

\bibitem[Zhao et~al.(2020)Zhao, Xiao, Gan, Zhang, and Xia]{zhao2020maintaining}
Bowen Zhao, Xi Xiao, Guojun Gan, Bin Zhang, and Shu-Tao Xia.
\newblock Maintaining discrimination and fairness in class incremental learning.
\newblock In \emph{CVPR}, 2020.

\bibitem[Zhou et~al.(2023{\natexlab{a}})Zhou, Ye, Zhan, and Liu]{zhou2023revisiting}
Da-Wei Zhou, Han-Jia Ye, De-Chuan Zhan, and Ziwei Liu.
\newblock Revisiting class-incremental learning with pre-trained models: Generalizability and adaptivity are all you need, 2023{\natexlab{a}}.

\bibitem[Zhou et~al.(2023{\natexlab{b}})Zhou, Zhang, Ning, Ye, Zhan, and Liu]{zhou2023learning}
Da-Wei Zhou, Yuanhan Zhang, Jingyi Ning, Han-Jia Ye, De-Chuan Zhan, and Ziwei Liu.
\newblock Learning without forgetting for vision-language models.
\newblock \emph{arXiv preprint arXiv:2305.19270}, 2023{\natexlab{b}}.

\bibitem[Zhu et~al.(2023)Zhu, Chen, Shen, Li, and Elhoseiny]{zhu2023minigpt}
Deyao Zhu, Jun Chen, Xiaoqian Shen, Xiang Li, and Mohamed Elhoseiny.
\newblock Minigpt-4: Enhancing vision-language understanding with advanced large language models.
\newblock \emph{arXiv preprint arXiv:2304.10592}, 2023.

\bibitem[Zhu et~al.(2021)Zhu, Zhang, Wang, Yin, and Liu]{zhu2021prototype}
Fei Zhu, Xu-Yao Zhang, Chuang Wang, Fei Yin, and Cheng-Lin Liu.
\newblock Prototype augmentation and self-supervision for incremental learning.
\newblock In \emph{Proceedings of the IEEE/CVF Conference on Computer Vision and Pattern Recognition}, pages 5871--5880, 2021.

\end{thebibliography}
